\documentclass[sigconf]{acmart}
\AtBeginDocument{%
  }

\setcopyright{acmlicensed}
\copyrightyear{2026}
\acmYear{2026}
\acmDOI{XXXXXXX.XXXXXXX}
\acmConference[Conference acronym ’XX]{Make sure to enter the correct conference title from your rights confirmation email}
{}{Woodstock, NY}
\acmISBN{978-1-4503-XXXX-X/2026/04}

\usepackage{graphicx}
\usepackage{subcaption}    
\usepackage{multirow}
\usepackage{enumitem}
\usepackage{amsmath}
\usepackage{amsthm}
\usepackage{mathtools}
\usepackage{wrapfig}
\usepackage{bbm}
\usepackage{colortbl}
\usepackage{nicematrix}
\usepackage{tabularx}
\usepackage{booktabs}       
\usepackage{microtype}      

\newcolumntype{b}{X}
\newcolumntype{s}{>{\hsize=.2\hsize}X}

\newlength\savewidth
\newcommand\shline{\noalign{\global\savewidth\arrayrulewidth
                            \global\arrayrulewidth 1.5pt}%
                   \hline
                   \noalign{\global\arrayrulewidth\savewidth}
                   }

\begin{document}

\title{FiCoTS: Fine-to-Coarse LLM-Enhanced Hierarchical Cross-Modality Interaction for Time Series Forecasting}

\author{Yafei Lyu}
\affiliation{%
  \institution{1. School of Advanced Interdisciplinary Sciences, University of Chinese Academy Sciences}
  \institution{2. MAIS, Institute of Automation, Chinese Academy of Science}    
  \city{Beijing}
    \country{China}} 
\email{lvyafei2025@ia.ac.cn}

\author{Hao Zhou}
\affiliation{%
  \institution{Great Bay University}
  \city{Dongguan}
  \country{China}}
\email{hao.zhou@hrbeu.edu.cn}

\author{Lu Zhang}
\affiliation{%
  \institution{MAIS, Institute of Automation, Chinese Academy of Science}
  \city{Beijing}
  \country{China}}
  \email{lu.zhang@ia.ac.c}

\author{Xu Yang}
\affiliation{%
  \institution{MAIS, Institute of Automation, Chinese Academy of Science}
  \city{Beijing}
  \country{China}}
\email{xu.yang@ia.ac.cn}

\author{Zhiyong Liu}
\affiliation{%
  \institution{MAIS, Institute of Automation, Chinese Academy of Science}
  \city{Beijing}
  \country{China}}
\email{zhiyong.liu@ia.ac.cn}

\renewcommand{\shortauthors}{Yafei Lyu et al.}

\begin{abstract}
Time series forecasting is central to data analysis and web technologies. The recent success of Large Language Models (LLMs) offers significant potential for this field, especially from the cross-modality aspect. Most methods adopt an LLM-as-Predictor paradigm, using LLM as the forecasting backbone and designing modality alignment mechanisms to enable LLM to understand time series data. However, the semantic information in the two modalities of time series and text differs significantly, making it challenging for LLM to fully understand time series data. To mitigate this challenge, our work follows an LLM-as-Enhancer paradigm to fully utilize the advantage of LLM in text understanding, where LLM is only used to encode text modality to complement time series modality. Based on this paradigm, we propose FiCoTS, an LLM-enhanced fine-to-coarse framework for multimodal time series forecasting. Specifically, the framework facilitates progressive cross-modality interaction by three levels in a fine-to-coarse scheme: First, in the token-level modality alignment module, a dynamic heterogeneous graph is constructed to filter noise and align time series patches with text tokens; Second, in the feature-level modality interaction module, a global cross-attention mechanism is introduced to enable each time series variable to connect with relevant textual contexts; Third, in the decision-level modality fusion module, we design a gated network to adaptively fuse the results of the two modalities for robust predictions. These three modules work synergistically to let the two modalities interact comprehensively across three semantic levels, enabling textual information to effectively support temporal prediction. Extensive experiments on seven real-world benchmarks demonstrate that our model achieves state-of-the-art performance. The codes will be released publicly.
\end{abstract}



\received{XX 2026}
\received[revised]{XX 2026}
\received[accepted]{XX 2026}

\maketitle

\section{Introduction}
Time series forecasting is a fundamental tool for various sequential data analysis tasks and lays the foundation for a variety of web services such as online weather forecasting, personalized content recommendation, etc.

Uncovering complex temporal dependency and making accurate predictions is crucial for time series forecasting. Traditional statistical methods, such as ARIMA \cite{anderson1976time} and Prophet \cite{taylor2018forecasting}, attempt to capture linear relationships but suffer from modeling complex non-linear time series patterns. Deep learning methods, including Recurrent Neural Networks (RNNs) \cite{qin2017dual}, Long Short-Term Memory Networks (LSTMs) \cite{lai2018modeling}, Convolutional Neural Networks (CNNs) \cite{wu2022timesnet}, Multi-Layer Perceptrons (MLPs) \cite{zeng2023transformers}, and Transformer-based architectures \cite{vaswani2017attention,wu2021autoformer,zhou2022fedformer,zhou2021informer,woo2022etsformer,nie2022time,liu2023itransformer}, show strong capability of capturing intricate non-linear and long-range dependencies, achieving state-of-the-art performance across various benchmarks. 

Despite their promising performance, the predictions of these models typically rely on historical numerical data with rare consideration of additional textual information, which to some extent hinders the model robustness and generalization ability. Recently, with the development of Large Language Models (LLMs) \cite{devlin2019bert,radford2019language,achiam2023gpt}, LLM-based time series forecasting methods have gained extensive attention \cite{jin2024position,liang2024foundation}. Existing methods can be categorized into two paradigms based on the role of LLM. The first paradigm is known as LLM-as-Predictor, which, inspired by the success of large language learning in the fields of for example natural language process and computer vision, directly constructs large models for time series. Technically, due to the data scarcity and training consumption of constructing a large model for time series from scratch \cite{garza2023timegpt,goswami2024moment,liu2024timer,shi2024time}, most researchers regard a pretrained LLM as the forecasting backbone. By designing specialized tokenizers to align representations of time series and text, they assume that the LLM is able to fully understand time series data \cite{zhou2023one,jin2023time,pan2024s,liu2025calf,sun2023test,cao2023tempo}. But the consequence turns out to be that in most scenarios, even a simple traditional temporal model may even outperform LLM-as-Predictor models \cite{tan2024language,zheng2024revisited}. Besides, even though most parameters of LLM are frozen, the inference process remains computationally expensive. The second paradigm is LLM-as-Enhancer by adopting a dual-stream framework, in which LLM acts as the encoder of text modality to enhance the time series modality \cite{liu2025timecma,zhong2025time} instead of processing time series data directly. LLM processes text prompts to generate contextual features, which are subsequently fused with the temporal features from the time series modality. However, existing strategies often facilitate only a shallow, coarse-grained interaction between the two modalities (e.g., simple feature concatenation), failing to establish the fine-grained, dynamic interaction. 

In this paper, we propose FiCoTS, a novel fine-to-coarse LLM-enhanced hierarchical cross-modality interaction framework for time series forecasting. It facilitates comprehensive cross-modality interaction through a three-level scheme, including token-level, feature-level, and decision-level interactions. Unlike the coarse-to-fine scheme which is commonly used in vision-language multimodal tasks \cite{yu2021cofinet,chen2023cf}, we employ a fine-to-coarse approach motivated by the inherent characteristics of time series forecasting. Specifically, FiCoTS is composed of three modules: First, in the token-level modality alignment module, we construct a dynamic heterogeneous graph to connect time series patches with semantically relevant text tokens based on the similarity matrix, enabling fine-grained, noise-filtered interaction; Second, in the feature-level modality interaction module, a global cross-attention mechanism is adopted to allow each variable sequence to focus on relevant textual contexts; Third, in the decision-level modality fusion module, a gate-based fusion approach is designed to adaptively combine predictions from both modalities, achieving robust forecasting results. This three-level scheme forms a cohesive and comprehensive fusion pipeline, ensuring that the complementary features of both modalities are thoroughly exploited. In summary, the main contributions of this work are as follows:
\begin{itemize}
    \item We propose a novel fine-to-coarse LLM-enhanced hierarchical cross-modality interaction framework for time series forecasting, after summarizing the two mainstream LLM-based time series forecasting paradigms of LLM-as-Predictor and LLM-as-Enhancer, and analyzing the characteristics of fine-to-coarse scheme and coarse-to-fine scheme.
    \item We construct a heterogeneous graph to align related text tokens and time series patches at a fine-grained level, where a dynamic edge filtering mechanism is also designed to remove weakly connected text tokens and time patches to reduce noise.
    \item Extensive experiments on seven real-world benchmarks demonstrate that our proposed framework outperforms a wide range of state-of-the-art methods, validating its effectiveness and superiority.
\end{itemize}

\section{Related Work}

\subsection{Deep Learning-Based Time Series Forecasting}

Classical models like ARIMA \cite{anderson1976time} and Prophet \cite{taylor2018forecasting} analyze time series from a statistic perspective and struggle to capture nonlinear patterns of time series. Following the development of deep learning, a number of methods have been proposed to solve time series tasks, such as Recurrent Neural Networks (RNNs) \cite{qin2017dual}, Long Short-Term Memory Networks (LSTMs) \cite{lai2018modeling}, Convolutional Neural Networks (CNNs) \cite{wu2022timesnet}, and Multi-Layer Perceptrons (MLPs) \cite{zeng2023transformers}. After that, Transformer-based models attract much attention. PatchTST \cite{nie2022time} adopts a channel-independence approach and segments time series into a set of patches, using the Transformer encoder to capture dependency between patches. iTransformer \cite{liu2023itransformer} applies an inverted way and treats the whole sequence of each channel as a patch, modeling the dependency between variables using Transformer. Despite their promising performance, these models have limited parameters and highly rely on historical numerical data, which constrains their generalization ability and interpretability.

\begin{figure}[htbp]
    \centering
    \begin{subfigure}[b]{0.23\textwidth} 
        \centering
        \includegraphics[width=\textwidth]{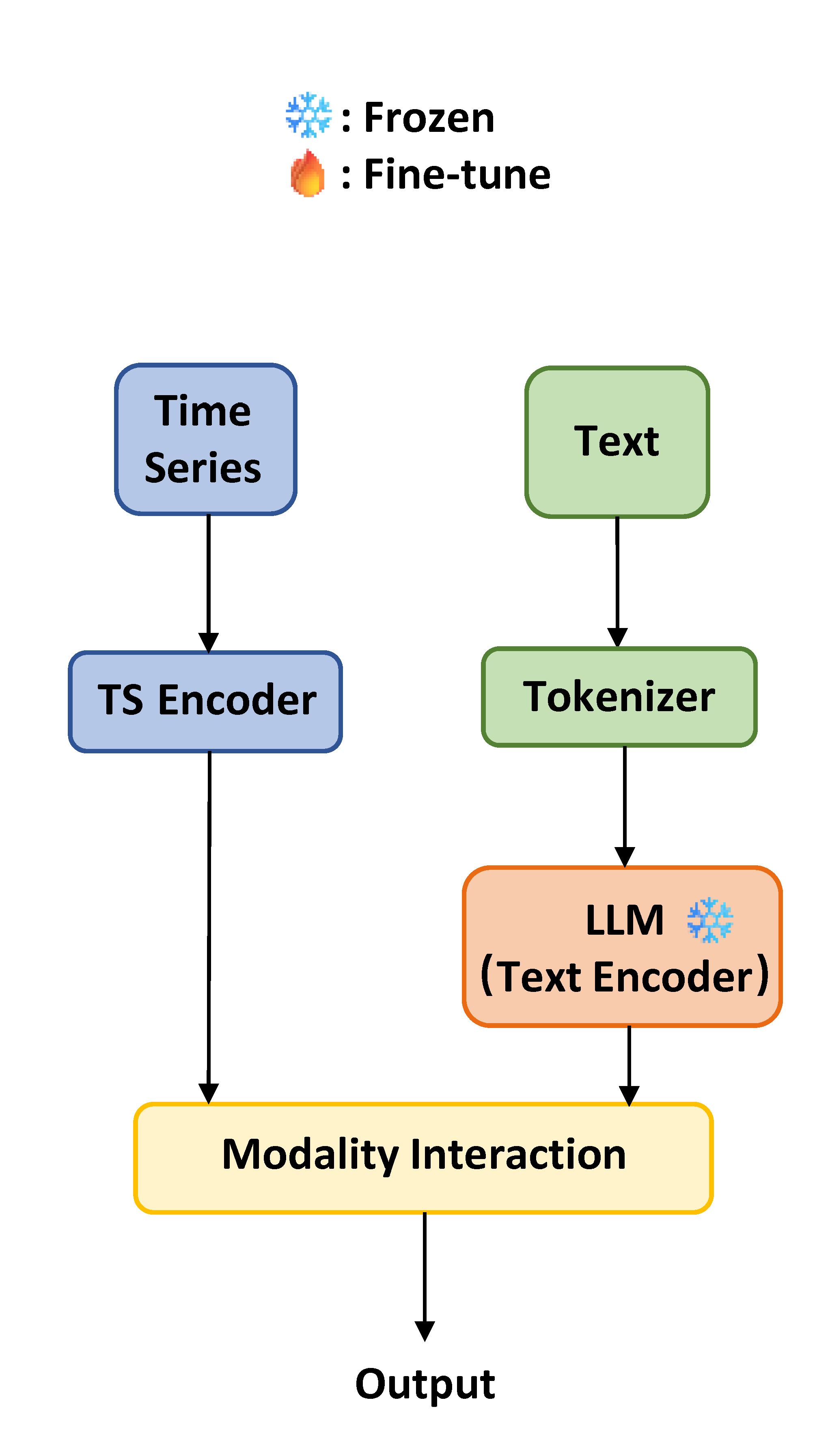}
        \caption*{(a) LLM-as-Enhancer}
        \label{fig:image1}
    \end{subfigure}
    \hfill
    \begin{subfigure}[b]{0.23\textwidth}   
        \centering
        \includegraphics[width=\textwidth]{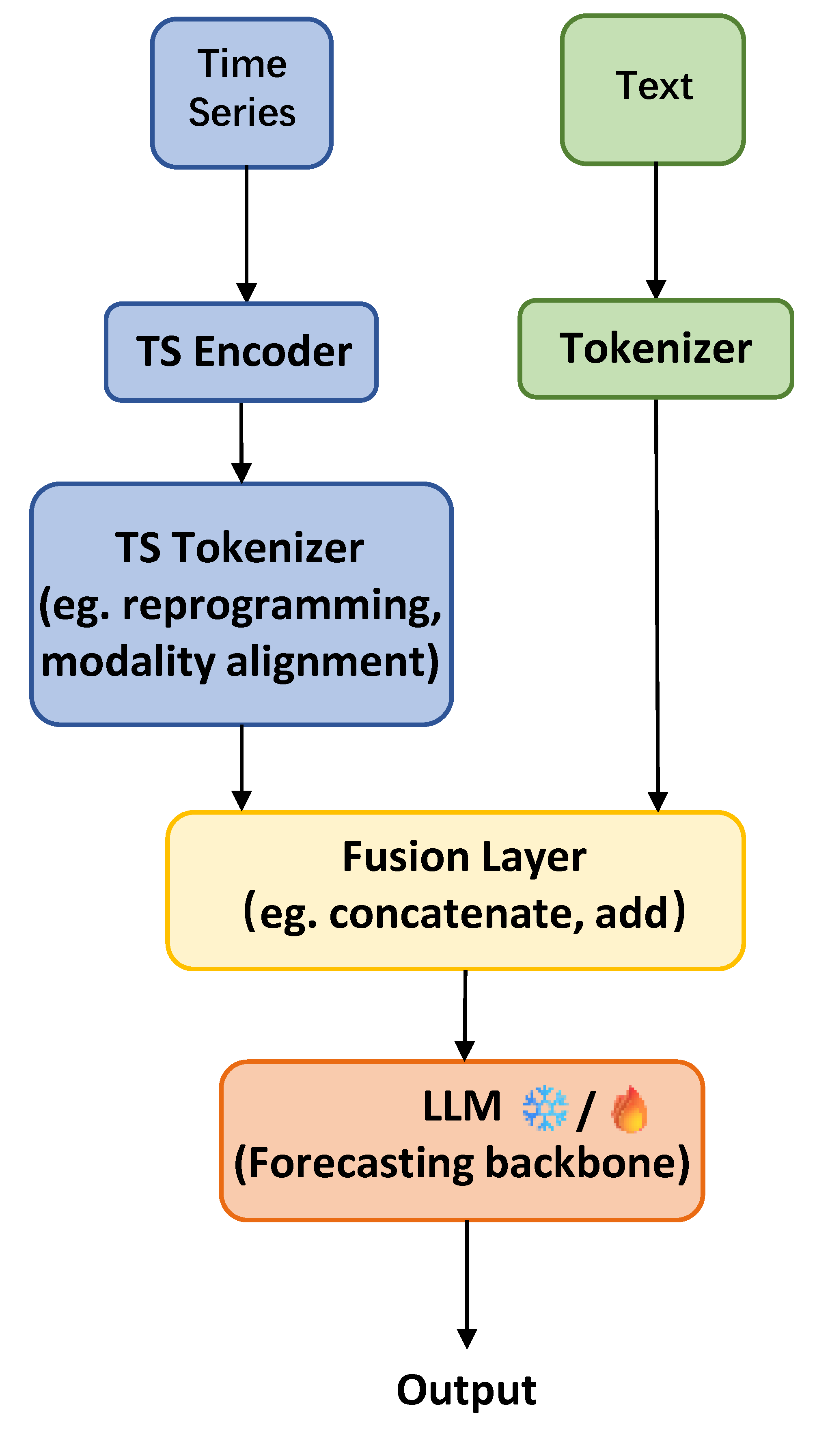}
        \caption*{(b) LLM-as-Predictor}
        \label{fig:image2}
    \end{subfigure}
    \caption{Comparison of the two LLM-based methods: (a) LLM-as-Enhancer. (b) LLM-as-Predictor.} 
\label{figure1}
\end{figure}

\begin{figure*}[t]
  \centering
  \includegraphics[width=\textwidth]{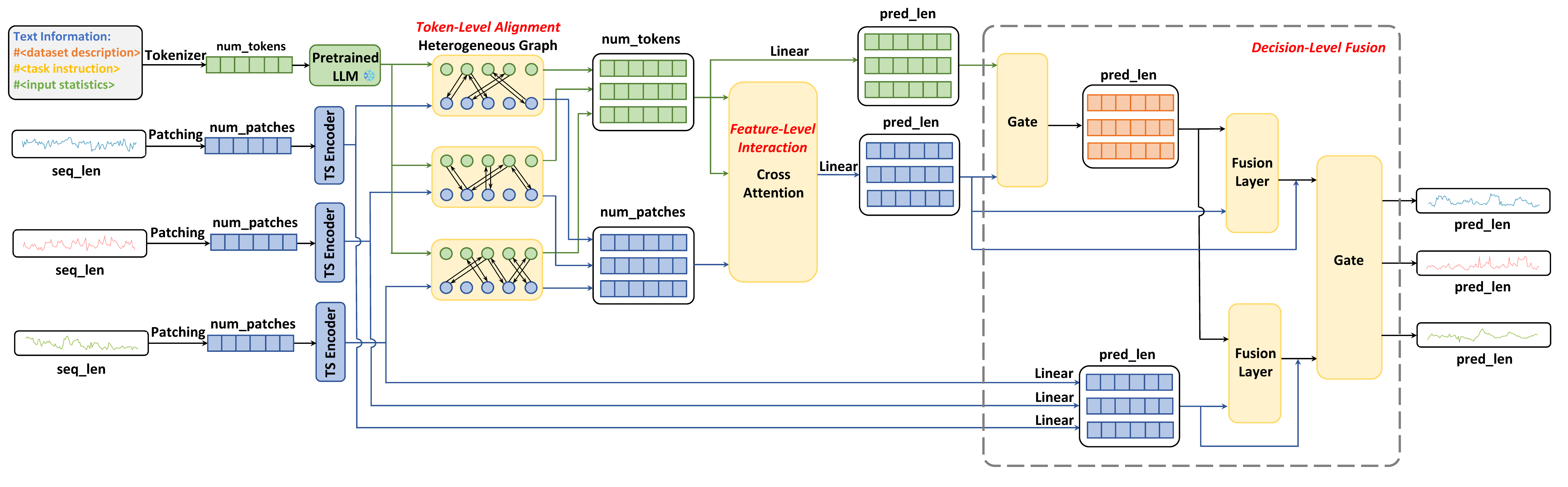}
  \caption{The model architecture of FiCoTS. The overall framework consists of three components: (1) Token-level alignment. (2) Feature-level interaction. (3) Decision-level fusion.}
\label{figure2}
\end{figure*}

\subsection{LLM-Based Time Series Forecasting}

Recently, LLM-based time series forecasting attracts significant attention \cite{jin2024position,liang2024foundation}. Owing to the scarcity of time series training data and the expensive computational cost, it is hard to pretrain and fine-tune a native large model for time series from scratch \cite{garza2023timegpt,goswami2024moment,liu2024timer,shi2024time}. Researchers tend to directly leverage the powerful pretrained large language model for forecasting. As shown in Figure \ref{figure1}, mainstream methods can be divided into two categories based on the role of LLM. 

\noindent\textbf{LLM-as-Predictor.} This method treats the pretrained LLM as the forecasting backbone and captures temporal dependency using Transformer encoder layers. In order to let LLM fully understand time series, designing modality alignment approaches and fine-tuning parameters of interface layers are the two common strategies. GPT4TS \cite{zhou2023one} first proposes the feasibility of applying LLM for time series analysis. It replaces the tokenizer of LLM by a linear embedding layer to transform time series patches into tokens. Time-LLM \cite{jin2023time} presents a patch reprogramming strategy to map time series patches into several text tokens and concatenates them with prompt as a prefix, enabling LLM to process time series data. TEST \cite{sun2023test} designs a three-wise contrast learning strategy to align the embedding spaces of time series and LLM. LLM4TS \cite{chang2025llm4ts} designs a two-stage fine-tuning strategy to adapt LLM to time series data. Despite their advances, Studies \cite{tan2024language,zheng2024revisited} indicate that LLM offers minimal advantages for time series analysis. Actually, simple models perform better in most scenarios. Besides, although most parameters of LLM were frozen, the forward propagation still brings a massive computational cost. 

\noindent\textbf{LLM-as-Enhancer.} Instead of regarding LLM as the forecasting backbone, this method pays attention to the significant impact of integrating multimodal information. It is explicit that the prediction of future sequences not only depends on the previous time series data, but it is also regulated by external factors, such as meteorological events, economic policies, or social dynamics. Similar to the mechanism of vision language model \cite{radford2021learning,li2022blip}, this method adopts a dual-tower architecture and treats text as an assistant modality, which carries useful external information for time series tasks. LLM is utilized to encode text data to complement time series modality, instead of processing time series directly. TimeCMA \cite{liu2025timecma} presents a channel-wise cross-modality alignment module to aggregate two modalities. TimeVLM \cite{zhong2025time} proposes a three-stream framework leveraging VLMs to process time series, image, and text. LLM-as-Enhancer paradigm makes it possible for several modalities to fully interact and complement each other for the forecasting task. Besides, it effectively saves cost computation and has stronger interpretability compared with LLM-as-Predictor method. However, LLM-as-Predictor method utilizes LLM to enable interaction between time series and text data, while LLM-as-Enhancer method requires carefully designed mechanisms for effective cross-modality interaction. Nevertheless, most existing approaches rely on simplistic fusion strategies like feature concatenation or weighted averaging, failing to establish fine-grained, hierarchical interactions between the two modalities. Our work presents a comprehensive cross-modality framework to facilitate progressive modality interaction and fully develop the capability of text modality for time series forecasting.

\section{Preliminaries}

\subsection{Problem Definition}

Given the historical multivariate time series data $\boldsymbol{X_{T}} = \{\boldsymbol{x}_{1},\ldots,\boldsymbol{x}_{T}\} \in \mathbb{R}^{T\times N}$ of $N$ variables across $T$ time steps, the goal of time series forecasting task is to learn a mapping function ${f}$ to predict the next $M$ time steps $\hat{\boldsymbol{X}}_{M} = f(\boldsymbol{X_{T}})= \{\hat{\boldsymbol{x}}_{T+1},\ldots,\hat{\boldsymbol{x}}_{T+M}\} \in \mathbb{R}^{M \times N}$. 

\subsection{Model Overview}

To effectively leverage the complementary information of text modality and empower the potential of LLM for time series, we present FiCoTS, a fine-to-coarse cross-modality interaction framework for time series forecasting. The overall architecture is illustrated in Figure \ref{figure2}. Firstly, both time series data and text data are processed by their own encoders. Here, we adopt LLM as the text encoder. Then, a dynamic heterogeneous graph is constructed to align the representation of time series patches and text tokens. Afterward, we perform cross-attention to let the two modalities globally interact at the feature-level. Finally, at the decision-level, we employ an adaptive gated mechanism, which combines the enhanced temporal features with contextualized text representations, ensuring robust prediction through dynamic modality weights. 

\section{Methodology}

Existing LLM-enhanced time series forecasting models face the challenge of inadequate cross-modality interaction. To address this issue, we present FiCoTS, a fine-to-coarse LLM-enhanced multimodal approach for time series forecasting. As illustrated in Figure \ref{figure2}, the overall architecture of FiCoTS mainly contains three components: token-level modality alignment, feature-level modality interaction, and decision-level modality fusion.

\subsection{Fine-to-Coarse or Coarse-to-Fine}
Instead of choosing a coarse-to-fine cross-modality interaction scheme, which is commonly adopted in vision-language tasks, we develop a fine-to-coarse scheme for the time series task. This design is motivated by the fundamental differences between multimodal time series forecasting and visual-language multimodal tasks.

\noindent\textbf{Data Structure and Task Characteristic.}
 Firstly, time series data has a 1-dimensional continuous structure. It is inherently dynamic and casual; the value of a previous point can significantly influence the next point. Local, fine-grained fluctuations are critical for forecasting. Therefore, it is crucial to capture the precise temporal dynamics and transient anomalies as a priority, which might be over-smoothed or lost in a coarse, global cross-modality interaction module. In contrast, visual information has a 2-dimensional spatial structure. Vision-language multimodal tasks like image captioning focus on visual understanding and textual description \cite{radford2021learning,li2022blip,ramesh2021zero}. They often begin with a coarse, global module to assess the overall situation. The filtered features are then refined with a fine-grained module for detailed information \cite{yu2021cofinet,chen2023cf,dou2022coarse,lu2024coarse}. 
 
\noindent\textbf{The Role of Text Modality.} In vision-language multimodal tasks, text modality is an indispensable component and often has explicit, fine-grained correspondence with a specific object in vision modality, naturally supporting a coarse-to-fine process which aligns global concepts before refining local details \cite{yu2021cofinet,chen2023cf,dou2022coarse,lu2024coarse}. Conversely, in time series forecasting tasks, the text serves as auxiliary information to enhance and complement the primary temporal data. Besides, there is no strict semantic connection between text content and time series data. While time series lack semantic information, it is unreasonable to establish a global interaction between text and time series at first. Therefore, starting with a fine-grained module allows time series patches to flexibly connect with relevant text tokens, endowing time series with fine-grained semantic information. On this basis, the following coarse-grained module can effectively achieve global cross-modality interaction, further enhancing the connection between two modalities. 

\subsection{Dual Stream Encoding}
\textbf{Time Series Encoding.} Given the historical multivariate time series $\boldsymbol{X_{T}}$ of $T$ time steps, we apply patching to segment time series sequence. Each input variable ${\boldsymbol{X}^{n}_{T}} \in \mathbb{R}^{T}, n=\left\{1,\ldots,N\right\}$ is divided into several overlapped or non-overlapped patches ${\boldsymbol{X}^{n}_{P}} \in \mathbb{R}^{P\times {L_{p}}}$. Here, $P = \lceil \frac{T - L_{p}}{S} \rceil + 1$ represents the patch numbers, where $L_{p}$ denotes the patch length and $S$ denotes the sliding stride. As demonstrated in \cite{zhang2025does}, the text modality provides a compensatory benefit when the time series encoder is not that strong. Therefore, we adopt a simple linear layer to embed time series patches into ${d_{m}}$ dimension, denoted as $\left\{\boldsymbol{x}^{n}_{i} \right\}_{i=1}^{P} \in \mathbb{R}^{P \times d_{m}}$.

\noindent\textbf{Text Information Encoding.} Followed LLM-as-Predictor method, we selected (1) dataset description, (2) task instruction and (3) input statistics as text prompt. The detailed textual prompt can be found in Appendix \ref{B}. We choose GPT2 as the text encoder to produce text tokens. Firstly, the text data is tokenized by GPT2 tokenizer and rejected into the dimension of pretrained LLM model, represented by ${\boldsymbol{P}^{n}} \in \mathbb{R}^{{N}_{p}\times {d_{m}}}$, where $N_{p}$ denotes the number of text tokens. Next, the tokenized text representation is processed by GPT2 layers to generate text embedding $\left\{\boldsymbol{p}^{n}_{j} \right\}_{j=1}^{{N}_{p}} \in \mathbb{R}^{{N}_{p} \times d_{m}}$.

\subsection{Token-Level Modality Alignment}
Graph structure can represent diverse data types and have great potential for cross-modality alignment. Therefore, we construct a heterogeneous graph to achieve token-level modality alignment. This heterogeneous graph not only explicitly presents the fine-grained associations between time series patches and text tokens, but also considers the inherent heterogeneity of each modality. 

\noindent\textbf{Dynamic Filtering Mechanism.} We compute the cosine similarity of time series tokens $\left\{\boldsymbol{x}^{n}_{i} \right\}_{i=1}^{P} \in \mathbb{R}^{P \times d_{m}}$ and text tokens $\left\{\boldsymbol{p}^{n}_{j} \right\}_{j=1}^{{N}_{p}} \in \mathbb{R}^{{N}_{p} \times d_{m}}$, getting the similarity matrix ${\boldsymbol{S}^{n}}\in \mathbb{R}^{{P}\times{{N}_{p}}}$. Aiming to eliminating unnecessary connections, we set a time-determined dynamic threshold for based on the mean ${\mu}_{i}$ and standard deviation ${\sigma}_{i}$ of temporal patch $\left\{\boldsymbol{x}^{n}_{i} \right\}_{i=1}^{P}$, where $\alpha$ controls the filtering sensitivity. The filtered similarity matrix is represented by 
\begin{align}
    {\boldsymbol{S}^{n}_{filtered}(i, j)} = \mathbb{I}\left( \boldsymbol{S}^{n}(i, j)\geq  {\mu}_{i} + {\alpha} \cdot {\sigma}_{i}\right).
\end{align}
The non-zero elements of ${\boldsymbol{S}^{n}_{filtered}}$ denote semantically relevant time series patches and text tokens. These temporal-textual pairs will be processed by the heterogeneous graph neural network later. This mechanism effectively filters temporal-textual pairs with weak connections, empowering time series with semantic information and mitigating noise. 

\noindent\textbf{Heterogeneous Graph Construction.} In order to effectively present the relationship between tokens of two modalities and achieve full interaction, a heterogeneous graph ${g}=\left(\mathcal{V}, \mathcal{E}, \Gamma\right)$ is constructed based the filtered temporal-text pairs. The node types are defined as $\Gamma=\left\{time, text\right\}$ to distinguish the two different modalities. The node set is denoted as $\mathcal{V}=\left\{\boldsymbol{x}^{n}_{i} \right\}_{i=1}^{P}\cup\left\{\boldsymbol{p}^{n}_{j} \right\}_{j=1}^{{N}_{p}}$.
Based on the similarity matrix ${\boldsymbol{S}^{n}_{filtered}}$, we constructed the edges $\mathcal{E}=\left\{\boldsymbol{x}^{n}_{i}\leftrightarrow\boldsymbol{p}^{n}_{j}|\boldsymbol{S}^{n}_{filtered}(i, j)>0\right\}$
between time series patchs and text tokens. Each edge is bidirectional to ensure relation-aware message passing between modalities.

\noindent\textbf{Heterogeneous Graph Learning.} To achieve token-level alignment, we adopt GraphSAGE operator to ensure effective information interaction:
\begin{align}
    N\left(\boldsymbol{x}^{n}_{i}\right)&=\left\{\boldsymbol{p}^{n}_{j}|\forall{j}, \boldsymbol{S}^{n}_{filtered}(i, j)>0\right\}, \\
    \boldsymbol{x}^n_{N(\boldsymbol{x}^{n}_{i})}&=\mathrm{Aggregate}({\boldsymbol{p}^{n}_{u}, \forall\boldsymbol{p}^{n}_{u}\in{N(\boldsymbol{x}^{n}_{i})}}),\\
    {\boldsymbol{x}^{n}_{i}}{’}&=\sigma(\boldsymbol{W}_{time}\cdot{\mathrm{Concat}[\boldsymbol{x}^{n}_{i}, \boldsymbol{x}^{n}_{N(\boldsymbol{x}^{n}_{i})}]}), i=1,\ldots,P,\\
    N\left(\boldsymbol{p}^{n}_{j}\right)&=\left\{\boldsymbol{x}^{n}_{i}|\forall{i}, \boldsymbol{S}^{n}_{filtered}(i, j)>0\right\},\\
    \boldsymbol{p}^n_{N(\boldsymbol{p}^{n}_{j})}&=\mathrm{Aggregate}({\boldsymbol{x}^{n}_{v}, \forall{\boldsymbol{x}^{n}_{v}}\in{N(\boldsymbol{p}^{n}_{j})}}),\\
    {\boldsymbol{p}^{n}_{j}}{’}&=\sigma(\boldsymbol{W}_{text}\cdot{\mathrm{Concat}[\boldsymbol{p}^{n}_{j}, \boldsymbol{p}^{n}_{N^(\boldsymbol{p}^{n}_{j})}]}), j=1,\ldots,N_{p}.
\end{align}
The aligned representations are denoted as $\boldsymbol{X}^n_{token}=\left\{{\boldsymbol{x}^{n}_{i}}{’}\right\}^{P}_{i=1}$ and $\boldsymbol{P}^n_{token}=\left\{{\boldsymbol{p}^{n}_{j}}{’}\right\}^{N_{p}}_{j=1}$.

\subsection{Feature-Level Modality Interaction}
\textbf{Cross-Attention Module.} After fine-grained token-level modality alignment, temporal tokens are endowed with semantic information. We calculate the mean value of temporal tokens to get global representation: 
\begin{align}
{\boldsymbol{\overline{{X}}}_{token}^{n}}=\mathrm{Mean}(\boldsymbol{X}^{n}_{token})\in \mathbb{R}^{d_{m}}.
\end{align}
Followed \cite{liu2025timecma}, we select the last token as the global text representation, as it includes the most comprehensive knowledge owing to the masked multi-head attention mechanism of LLM:
\begin{align}
{\boldsymbol{\overline{{P}}}_{token}^{n}}={\boldsymbol{p}^{n}_{N_{p}}}{’}\in \mathbb{R}^{d_{m}}.
\end{align}
Given the global representations ${\boldsymbol{\overline{{X}}}_{token}}\in \mathbb{R}^{d_{m} \times N}$ and ${\boldsymbol{\overline{{P}}}_{token}}\in \mathbb{R}^{d_{m} \times N}$, we adopt multi-head cross-attention (MCA) on the two modalities:
\begin{align}
\mathrm{MCA}(\boldsymbol{Q}, \boldsymbol{K}, \boldsymbol{V})&=\mathrm{Cat}(\boldsymbol{head}_{1}, \ldots, \boldsymbol{head}_{h})\boldsymbol{W}^{feature},\\
\boldsymbol{head}_{m}&=\mathrm{Softmax}\left(\frac{{\boldsymbol{QW}^{Q}_{m}(\boldsymbol{KW}^{K}_{m})^{\top}}}{\sqrt{d_{k}}}\right)\boldsymbol{VW}^{V}_{m},\\
\boldsymbol{Q}&=\boldsymbol{\overline{{X}}}_{token}\boldsymbol{W}^{Q},
\\\boldsymbol{K}&=\boldsymbol{\overline{{P}}}_{token}\boldsymbol{W}^{K},
\\
\boldsymbol{V}&=\boldsymbol{\overline{{P}}}_{token}\boldsymbol{W}^{V},
\end{align}
where time series modality serves as the query, and text modality serves as the key and value. This mechanism allows temporal feature to flexibly interact with relevant textual feature in the medium-grained feature-level, endowing time series representation with global semantic information. The final features are represented by ${\boldsymbol{X}_{feature}}$
and ${\boldsymbol{P}_{feature}}$:
\begin{align}
{\boldsymbol{X}_{feature}}&=\mathrm{MCA}(\boldsymbol{Q}, \boldsymbol{K}, \boldsymbol{V})\in \mathbb{R}^{{{d}_{m}}\times{N}},\\{\boldsymbol{P}_{feature}}&=\mathrm{LayerNorm}(\boldsymbol{\overline{{P}}}_{token})\in \mathbb{R}^{{{d}_{m}}\times{N}}.
\end{align}

\subsection{Decision-Level Modality Fusion}
\textbf{Adaptive Gate Fusion.} After the previous token-level and feature-level cross-modality interaction, the two modalities exchange information and generate comprehensive features. For the purpose of fully integrating both modalities and achieving precise prediction, we design an adaptive gated fusion module to effectively fuse the two modalities in decision-level. Firstly, the temporal features ${\boldsymbol{X}_{feature}}$ and the textual features ${\boldsymbol{P}_{feature}}$ are both projected into $M$ dimension, denoted as $\hat{\boldsymbol{X}}_{feature}\in \mathbb{R}^{{M}\times{N}}$ and $\hat{\boldsymbol{P}}_{feature}\in \mathbb{R}^{{M}\times{N}}$, where $M$ is the prediction length. Then the two modalities are aggregated by a gated fusion module, generating the fused multimodal feature:
\begin{align}
    \boldsymbol{G}_{gate}&=\sigma(\boldsymbol{W}_{gate}[\hat{\boldsymbol{X}}_{feature}; \hat{\boldsymbol{P}}_{feature}]+\boldsymbol{b}_{gate}),
        \\
\boldsymbol{F}_{m}&=\boldsymbol{G}_{gate}\odot\hat{\boldsymbol{X}}_{feature}+\left(1-\boldsymbol{G}_{gate}\right)\odot\hat{\boldsymbol{P}}_{feature}.
\end{align}
Considering that the original temporal patterns are somewhat mitigated after the preceding cross-modality interaction, we design a decision-level fusion module to adaptively combine predictions from both modalities. Specifically, this module contains two branches. Firstly, the temporal feature $\hat{\boldsymbol{X}}_{feature}$ and the fused multimodal feature $\boldsymbol{F}_{m}$ are aggregated by a linear fusion layer, after which the temporal feature is further added by a residual connection:
\begin{align}
\boldsymbol{X}_{decision,1}=\mathrm{Linear}(\boldsymbol{F}_{m})+\hat{\boldsymbol{X}}_{feature}.
\end{align}
Besides, the original time series patches, which carry valuable original time series features, are projected into the $M$ dimension  to get $\hat{\boldsymbol{X}}_{original}\in \mathbb{R}^{{M}\times{N}}$ . Then we combine $\hat{\boldsymbol{X}}_{original}$ and $\boldsymbol{F}_{m}$ in the same way of $\hat{\boldsymbol{X}}_{feature}$ and $\boldsymbol{F}_{m}$:
\begin{align}
\boldsymbol{X}_{decision,2}=\mathrm{Linear}(\boldsymbol{F}_{m})+\hat{\boldsymbol{X}}_{original}.
\end{align}
Ultimately, the two predictions are fused by a gated module to adaptively combine the two branches based on their importance. The final prediction is denoted as $\hat{\boldsymbol{X}}_{M}\in \mathbb{R}^{{M}\times{N}}$:
\begin{align}
&\boldsymbol{G}_{decision}=\sigma(\boldsymbol{W}_{decision}[\boldsymbol{X}_{decision,1}; \boldsymbol{X}_{decision,2}]+\boldsymbol{b}_{decision}),\\
&\hat{\boldsymbol{X}}_{M}=\boldsymbol{G}_{decision}\odot\boldsymbol{X}_{decision,1}+\left(1-\boldsymbol{G}_{decision}\right)\odot\boldsymbol{X}_{decision,2}.
\end{align}

In conclusion, the hierarchical integration from token-level alignment through feature-level refinement to decision-level fusion not only ensures comprehensive utilization of multimodal information, but also maintains the temporal integrity for accurate forecasting.

\subsection{Objective Function and Optimization}
We apply Mean Squared Error (MSE) as the forecasting loss:
\begin{align}
\mathcal{L} = \frac{1}{M} \sum_{m=1}^{M} || \boldsymbol{\hat{x}}_{m} - \boldsymbol{x}_{m} ||^{2},    
\end{align}
where $\boldsymbol{x}_{m}\in \mathbb{R}^{N}$ denotes the ground-truth value of a time step and $\hat{\boldsymbol{x}}_{m}$ denotes the model's prediction. The model is trained following an end-to-end paradigm based on the observed values of  the next $M$ time steps.

\section{Experiments}
\subsection{Experimental Setup}
\textbf{Datasets.} We conduct a comprehensive evaluation of the proposed FiCoTS model on seven real-world benchmark datasets: (1) \textbf{ETT} consists of four sub-datasets of ETTm1, ETTm2, ETTh1, and ETTh2, which include key indicators collected for monitoring the operation status of electricity transformers from July 2016 to July 2018. 
(2) \textbf{Electricity} records the hourly electricity consumption of 321 customers between 2012 and 2014.
(3) \textbf{Weather} contains 21 meteorological variables—such as air temperature, humidity, and rainfall—measured at 10-minute intervals throughout the year 2020.
(4) \textbf{Traffic} includes the hourly road occupancy recorded by sensors on San Francisco highways from 2015 to 2016.
As summarized in Appendix \ref{A1}, the time series datasets from different domains exhibit notable variations in characteristics such as the number of variables, sampling frequencies, and overall data size. The diversity of these datasets enables us to thoroughly validate the performance of the model. 

\noindent\textbf{Baselines.} We compare with the SOTA state-of-art time series forecasting models, including Transformer-based models like PatchTST \cite{nie2022time}, iTransformer \cite{liu2023itransformer}, Autoformer \cite{wu2021autoformer}, FEDformer \cite{zhou2022fedformer} and ETSformer \cite{woo2022etsformer}; and non-Transformer-based models like DLinear \cite{zeng2023transformers} and TimesNet \cite{wu2022timesnet}; We also select very recent LLM-as-Predictor models, including GPT4TS \cite{zhou2023one}, Time-LLM \cite{jin2023time} and S$^{2}$IP-LLM \cite{pan2024s}. Besides, the LLM-as-Enhancer model TimeVLM \cite{zhong2025time} is also used for comparison.

\begin{table*}[t]
    \centering
    \small
    \tabcolsep=0.8mm
    \caption{Long-term forecasting results. The input sequence length is set to 512. The predictive lengths are set to $\{96, 192, 336, 720\}$. We bold the best performance and underline the second best performance.}
    \vspace{-1.5em}
    \resizebox{\textwidth}{!}{
    \begin{tabular}{c|cccccccccc||cccc||cccccccccc}
    \shline
    \multicolumn{1}{c}{\multirow{3}{*}{Method}} & \multicolumn{10}{c||}{\textbf{\textit{LLM-based}}} & \multicolumn{4}{c||}{\textbf{\textit{Non-Transformer-based}}} & \multicolumn{10}{c}{\textbf{\textit{Transformer-based}}} \\ 
    \cline{2-25}
    \multicolumn{1}{c}{} & \multicolumn{2}{c}{FiCoTS} & \multicolumn{2}{c}{TimeVLM} & \multicolumn{2}{c}{S$^{2}$IP-LLM} & \multicolumn{2}{c}{TimeLLM} & \multicolumn{2}{c||}{GPT4TS} & \multicolumn{2}{c}{TimesNet} & \multicolumn{2}{c||}{DLinear} & \multicolumn{2}{c}{PatchTST} & \multicolumn{2}{c}{iTransformer} & \multicolumn{2}{c}{FEDformer} & \multicolumn{2}{c}{Autoformer} & \multicolumn{2}{c}{ETSformer} \\ 
    \cline{2-25} 
    \multicolumn{1}{c}{} & MSE & MAE & MSE & MAE & MSE & MAE & MSE & MAE & MSE & MAE & MSE & MAE & MSE & MAE & MSE & MAE & MSE & MAE & MSE & MAE & MSE & MAE & MSE & MAE\\
    \hline 
    ETTh1 & \textbf{0.398} & \textbf{0.419} & \underline{0.405} & \underline{0.420} & 0.418 & 0.436 & 0.426 & 0.435 & 0.427 & 0.436 & 0.458 & 0.450 & 0.422 & 0.437 & 0.413 & 0.430 & 0.451 & 0.462 & 0.440 & 0.460 & 0.496 & 0.487 & 0.542 & 0.510\\
    \hline 
    ETTh2 & \underline{0.335} & \underline{0.385} & 0.341 & 0.391 & 0.355 & 0.399 & 0.361 & 0.398 & 0.354 & 0.394 & 0.414 & 0.427 & 0.431 & 0.446 & \textbf{0.330} & \textbf{0.379} & 0.382 & 0.414 & 0.437 & 0.449 & 0.450 & 0.459 & 0.439 & 0.452 \\
    \hline 
    ETTm1 & \textbf{0.346} & \textbf{0.373} & \underline{0.350} & 0.377 & 0.346 & 0.382 & 0.354 & 0.384 & 0.352 & 0.383 & 0.400 & 0.406 & 0.357 & 0.378 & 0.351 & 0.380 & 0.370 & 0.399 & 0.448 & 0.452 & 0.588 & 0.517 & 0.429 & 0.425 \\
    \hline 
    ETTm2 & \underline{0.251} & \underline{0.312} & \textbf{0.248} & \textbf{0.311} & 0.262 & 0.326 & 0.275 & 0.334 & 0.266 & 0.326 & 0.291 & 0.333 & 0.267 & 0.333 & 0.255 & 0.315 & 0.272 & 0.331 & 0.305 & 0.349 & 0.327 & 0.371 & 0.293 & 0.342 \\
    \hline 
    Weather & \textbf{0.224} & \textbf{0.262} & \textbf{0.224} & \underline{0.263} & 0.228 & 0.265 & 0.237 & 0.269 & 0.237 & 0.270 & 0.259 & 0.287 & 0.248 & 0.300 & 0.225 & 0.264 & 0.304 & 0.335 & 0.309 & 0.360 & 0.338 & 0.382 & 0.271 & 0.334 \\
    \hline 
    Electricity & \textbf{0.161} & \underline{0.254} & 0.172 & 0.273 & 0.166 & 0.262 & 0.167 & 0.264 & 0.167 & 0.263 & 0.192 & 0.295 & 0.166 & 0.263 & \textbf{0.161} & \textbf{0.252} & 0.203 & 0.298 & 0.214 & 0.327 & 0.227 & 0.338 & 0.208 & 0.323 \\
    \hline 
    Traffic & 0.403 & \underline{0.274} & 0.419 & 0.303 & 0.405 & 0.286 & 0.407 & 0.289 & 0.414 & 0.294 & 0.620 & 0.336 & \underline{0.433} & 0.295 & \underline{0.390} & \textbf{0.263} & \textbf{0.389} & 0.295 & 0.610 & 0.376 & 0.628 & 0.379 & 0.621 & 0.396 \\
    \shline
    \end{tabular}
    } 
\label{table1}
\end{table*}

\begin{table*}[h]
    \centering
    \small
    \tabcolsep=0.8mm
    \caption{Few-shot forecasting results on 10\% training data. The input sequence length is set to 512. The predictive lengths are set to $\{96, 192, 336, 720\}$. We bold the best performance and underline the second best performance.}
    \vspace{-1.5em}
    \resizebox{\textwidth}{!}{
    \begin{tabular}{c|cccccccccc||cccc||cccccccccc}
    \shline
    \multicolumn{1}{c}{\multirow{3}{*}{Method}} & \multicolumn{10}{c||}{\textbf{\textit{LLM-based}}} & \multicolumn{4}{c||}{\textbf{\textit{Non-Transformer-based}}} & \multicolumn{10}{c}{\textbf{\textit{Transformer-based}}} \\ 
    \cline{2-25}
    \multicolumn{1}{c}{} & \multicolumn{2}{c}{FiCoTS} & \multicolumn{2}{c}{TimeVLM} & \multicolumn{2}{c}{S$^{2}$IP-LLM} & \multicolumn{2}{c}{TimeLLM} & \multicolumn{2}{c||}{GPT4TS} & \multicolumn{2}{c}{TimesNet} & \multicolumn{2}{c||}{DLinear} & \multicolumn{2}{c}{PatchTST} & \multicolumn{2}{c}{iTransformer} & \multicolumn{2}{c}{FEDformer} & \multicolumn{2}{c}{Autoformer} & \multicolumn{2}{c}{ETSformer} \\ 
    \cline{2-25} 
    \multicolumn{1}{c}{} & MSE & MAE & MSE & MAE & MSE & MAE & MSE & MAE & MSE & MAE & MSE & MAE & MSE & MAE & MSE & MAE & MSE & MAE & MSE & MAE & MSE & MAE & MSE & MAE\\
    \hline 
    ETTh1 & \textbf{0.421} & \textbf{0.438} & \underline{0.431} & \underline{0.442} & 0.593 & 0.529 & 0.785 & 0.553 & 0.590 & 0.525 & 0.869 & 0.628 & 0.691 & 0.600 & 0.633 & 0.542 & 0.910 & 0.860 & 0.639 & 0.561 & 0.702 & 0.596  & 1.180 & 0.834\\
    \hline 
    ETTh2 & \textbf{0.341} & \textbf{0.389} & \underline{0.361} & \underline{0.405} & 0.419 & 0.439 & 0.424 & 0.441 & 0.397 & 0.421 & 0.479 & 0.465 & 0.605 & 0.538 & 0.415 & 0.431 & 0.489 & 0.483 & 0.466 & 0.475 & 0.488 & 0.499 & 0.894 & 0.713 \\
    \hline 
    ETTm1 & \textbf{0.361} & \textbf{0.382} & \underline{0.364} & \underline{0.385} & 0.455 & 0.446 & 0.477 & 0.451 & 0.472 & 0.450 & 0.717 & 0.561 & 0.400 & 0.417 & 0.526 & 0.476 & 0.784 & 0.596 & 0.730 & 0.592 & 0.796 & 0.620 & 1.125 & 0.782 \\
    \hline 
    ETTm2 & \textbf{0.257} & \textbf{0.317} & \underline{0.262} & \underline{0.323} & 0.296 & 0.342 & 0.307 & 0.348 & 0.308 & 0.346 & 0.344 & 0.372 & 0.399 & 0.426 & 0.314 & 0.352 & 0.356 & 0.388 & 0.381 & 0.404 & 0.388 & 0.433 & 0.534 & 0.547 \\
    \shline
    \end{tabular}
    } 
\label{table2}
\end{table*}

\begin{table*}[h]
    \centering
    \small
    \tabcolsep=0.8mm
    \caption{Few-shot forecasting results on 5\% training data. The input sequence length is set to 512. The predictive lengths are set to $\{96, 192, 336, 720\}$. We bold the best performance and underline the second best performance.}
    \vspace{-1.5em}
    \resizebox{\textwidth}{!}{
    \begin{tabular}{c|cccccccccc||cccc||cccccccccc}
    \shline
    \multicolumn{1}{c}{\multirow{3}{*}{Method}} & \multicolumn{10}{c||}{\textbf{\textit{LLM-based}}} & \multicolumn{4}{c||}{\textbf{\textit{Non-Transformer-based}}} & \multicolumn{10}{c}{\textbf{\textit{Transformer-based}}} \\ 
    \cline{2-25}
    \multicolumn{1}{c}{} & \multicolumn{2}{c}{FiCoTS} & \multicolumn{2}{c}{TimeVLM} & \multicolumn{2}{c}{S$^{2}$IP-LLM} & \multicolumn{2}{c}{TimeLLM} & \multicolumn{2}{c||}{GPT4TS} & \multicolumn{2}{c}{TimesNet} & \multicolumn{2}{c||}{DLinear} & \multicolumn{2}{c}{PatchTST} & \multicolumn{2}{c}{iTransformer} & \multicolumn{2}{c}{FEDformer} & \multicolumn{2}{c}{Autoformer} & \multicolumn{2}{c}{ETSformer} \\ 
    \cline{2-25} 
    \multicolumn{1}{c}{} & MSE & MAE & MSE & MAE & MSE & MAE & MSE & MAE & MSE & MAE & MSE & MAE & MSE & MAE & MSE & MAE & MSE & MAE & MSE & MAE & MSE & MAE & MSE & MAE\\
    \hline 
    ETTh1 & \textbf{0.426} & \textbf{0.440} & \underline{0.442} & \underline{0.453} & 0.650 & 0.550 & 0.648 & 0.549 & 0.681 & 0.560 & 0.925 & 0.647 & 0.750 & 0.611 & 0.694 & 0.569 & 1.070 & 0.710 & 0.658 & 0.562 & 0.722 & 0.598 & 1.189 & 0.839\\
    \hline 
    ETTh2 & \textbf{0.337} & \textbf{0.382} & \underline{0.354} & \underline{0.402} & 0.380 & 0.413 & 0.398 & 0.426 & 0.400 & 0.433 & 0.439 & 0.448 & 0.694 & 0.577 & 0.827 & 0.615 & 0.488 & 0.475 & 0.463 & 0.454 & 0.441 & 0.457 & 0.809 & 0.681 \\
    \hline 
    ETTm1 & \textbf{0.361} & \textbf{0.382} & \underline{0.364} & \underline{0.385} & 0.455 & 0.446 & 0.477 & 0.451 & 0.472 & 0.450 & 0.717 & 0.561 & 0.400 & 0.417 & 0.526 & 0.476 & 0.784 & 0.596 & 0.730 & 0.592 & 0.796 & 0.620 & 1.125 & 0.782 \\
    \hline 
    ETTm2 & \textbf{0.257} & \textbf{0.317} & \underline{0.262} & \underline{0.323} & 0.296 & 0.342 & 0.307 & 0.348 & 0.308 & 0.346 & 0.344 & 0.372 & 0.399 & 0.426 & 0.314 & 0.352 & 0.356 & 0.388 & 0.381 & 0.404 & 0.388 & 0.433 & 0.534 & 0.547 \\
    \shline
    \end{tabular}
    } 
\label{table3}
\end{table*}

\noindent\textbf{Implementation Details.} Consistent with LLM-based time series forecasting models, we use the pretrained GPT-2 as the default pretrained large language model with its parameters frozen. Besides, we leverage Adam as an optimizer and set the initial learning rate to be 0.001. The batch size is 32 for all models to guarantee fair comparison. In addition, we set the training epochs to 20 and employ an early stopping strategy to avoid overfitting. All experiments are conducted on NVIDIA RTX 4090 GPUs using the Pytorch deep learning framework.

\subsection{Long-Term Forecasting}

\textbf{Setup.} For long-term forecasting task, we conduct experiments on seven commonly used datasets (ETTh1, ETTh2, ETTm1, ETTm2, Weather, Electricity and Traffic). Following LLM-based models, we set the input sequence length to 512 time steps. The prediction length ranges from $\{96, 192, 336, 720\}$. We apply Mean Squared Error (MSE) and Mean Absolute Error (MAE) as evaluation metrics.

\noindent\textbf{Results.} Table \ref{table1} presents the performance of FiCoTS compared with baselines on seven real-world datasets. The detailed results are presented in Appendix \ref{C}. We can observe that LLM-based models surpass traditional temporal models, which illustrates the strong potential of leveraging LLM's potential capability for the time series forecasting task. Besides, LLM-as-Enhancer models perform better than LLM-as-Predictor models, revealing that treating LLM as an auxiliary is the better choice to apply LLM in the time series forecasting task.

\vspace{-0.2cm}
\subsection{Few-Shot Forecasting}
\textbf{Setup.} Considering that LLMs and multimodal models show outstanding capabilities in few-shot scenarios, we test the performance of FiCoTS with only 10\% or 5\% training data to evaluate the few-shot forecasting ability of FiCoTS. Experiments are conducted on ETT datasets. Other setups are consistent with the long-term forecasting task. 

\noindent \textbf{Results.} The results under 10\% and 5\% training data are presented in Table \ref{table2} and Table \ref{table3}, respectively. The detailed results can be found in Appendix \ref{D}. The results show that FiCoTS achieves consistent improvement over baselines, illustrating its strong few-shot capability. Moreover, we observe that FiCoTS outperforms LLM-as-Predictor models, which demonstrates that without using LLM as the forecasting backbone, LLM-as-Enhancer models can also leverage the strong few-shot ability of LLMs. 

\vspace{-0.2cm}
\subsection{Ablation Studies}
We conducted ablation studies to validate the necessity of the main components in FiCoTS. The results are provided in Table \ref{table4}.\\
\noindent \textbf{Ablation on Hierarchical Modality Interaction Components.} To evaluate the necessities of the three components in our fine-to-coarse modality interaction framework, we conducted ablation studies on ETTh1 and ETTm1 datasets. For the ETTh1 dataset, removing token-level, feature-level, and decision-level modules results in a performance reduction of 2.7\%, 0.7\% and 6.0\% in MSE and 2.1\%, 0.7\% and 5.0\% in MAE, respectively. For the ETTm1 dataset, removing these three components also causes 1.3\%, 0.8\% and 2.1\% performance decline in MAE. Moreover, for the decision-level, we conduct ablations on removing each branch respectively. In ETTh1 and ETTm1 datasets, the results indicate that removing the aligned temporal branch (branch1) causes a 3\% and 2\% performance drop in MSE. Similarly, the absence of the original temporal branch (branch2) results in a 2.7\% and 1.7\% performance degradation. The ablation results demonstrate the contribution of each module. Only through the synergistic interplay of token-level, feature-level and decision-level modules can optimal results be achieved.

\noindent \textbf{Ablation on Text Modality.} We also conduct an ablation study to demonstrate the importance of LLM in text modality. For ETTh1 and ETTm1 datasets, ablation of LLM results in 1.7\% and 1.1\% performance decline, which illustrates the significant role of LLM in processing text content and therefore interacting with time series modality.

\noindent \textbf{Ablation on the Construction of Intra-modality Edges.} To further explore the capability of the constructed heterogeneous neural network, we conduct an ablation study to investigate whether to connect intra-modality connections within individual modalities. The ablation result shows that adding intra-modality edges (i.e., connecting time steps to other time steps, and text tokens to other text tokens) leads to a degradation in performance. We attribute this result to the mismatch of sequence data and graph structure. Unlike image data which has an explicit spatial connection that can be well presented by a graph, time series and text data are sequential and causal. Enforcing graph connection on these types of data may lead to over-smoothing of temporal patterns and dilute important long-term dependencies.

\noindent \textbf{Ablation on Heterogeneous Graph or Homogeneous Graph.} To examine the advantage of heterogeneous graph for cross-modality alignment, we conduct an ablation study by replacing it with a homogeneous graph. The result shows a performance decline, demonstrating that compared with homogeneous graph, heterogeneous graph keeps the inherent heterogeneity of modalities, which leads to better modality alignment quality.

\noindent \textbf{Ablation on Types of GNN.} We conduct an ablation on GNN types to select an appropriate interaction mechanism. By replacing GraphSAGE with GCN, we got a slight performance decline, which illustrates that the advanced aggregate mechanism of GraphSAGE is more effective for cross-modality alignment.

\begin{table}[h]
    \centering
    \small
    \tabcolsep=0.7mm
    \caption{Ablation studies on ETTh1 and ETTm1 datasets for long-term forecasting.}
    \vspace{-1.5em}
    \resizebox{0.85\linewidth}{!}{
    \begin{tabular}{c|cc|cc}
    \shline
    \multirow{2}{*}{Variant} & \multicolumn{2}{c|}{ETTh1} & \multicolumn{2}{c}{ETTm1} \\ \cline{2-5} 
    & MSE & MAE & MSE & MAE \\
    \hline \hline
    FiCoTS & \textbf{0.398} & \textbf{0.419} & \textbf{0.346} & \textbf{0.373} \\
    w/o token-level & 0.409 & 0.428 & 0.357 & 0.378 \\
    w/o feature-level & \underline{0.401} & \underline{0.422} & \textbf{0.346} & {0.376} \\
    w/o decision-level & 0.423 & {0.441} & 0.350 & 0.381 \\
    w/o branch1 & 0.410 & 0.431 & 0.353 & 0.378 \\
    w/o  branch2 & 0.409 & 0.430 & 0.352 & 0.379 \\
    w/o  LLM & 0.405 & 0.424 & \underline{0.347} & \underline{0.375} \\
     w/o  intra-modality & 0.405 & 0.425 & 0.350 & 0.377 \\
     homogeneous & \underline{0.401} & \underline{0.422} & \underline{0.347} & 0.377 \\
     GCN & \underline{0.401} & 0.423 & 0.349 & 0.377 \\
    \shline
    \end{tabular}
    }
    \label{tab:ablation}
    \vspace{-1em}
\label{table4}
\end{table}

\subsection{Model Analysis}

\begin{figure}[h]
  \centering
  \includegraphics[width=\linewidth]{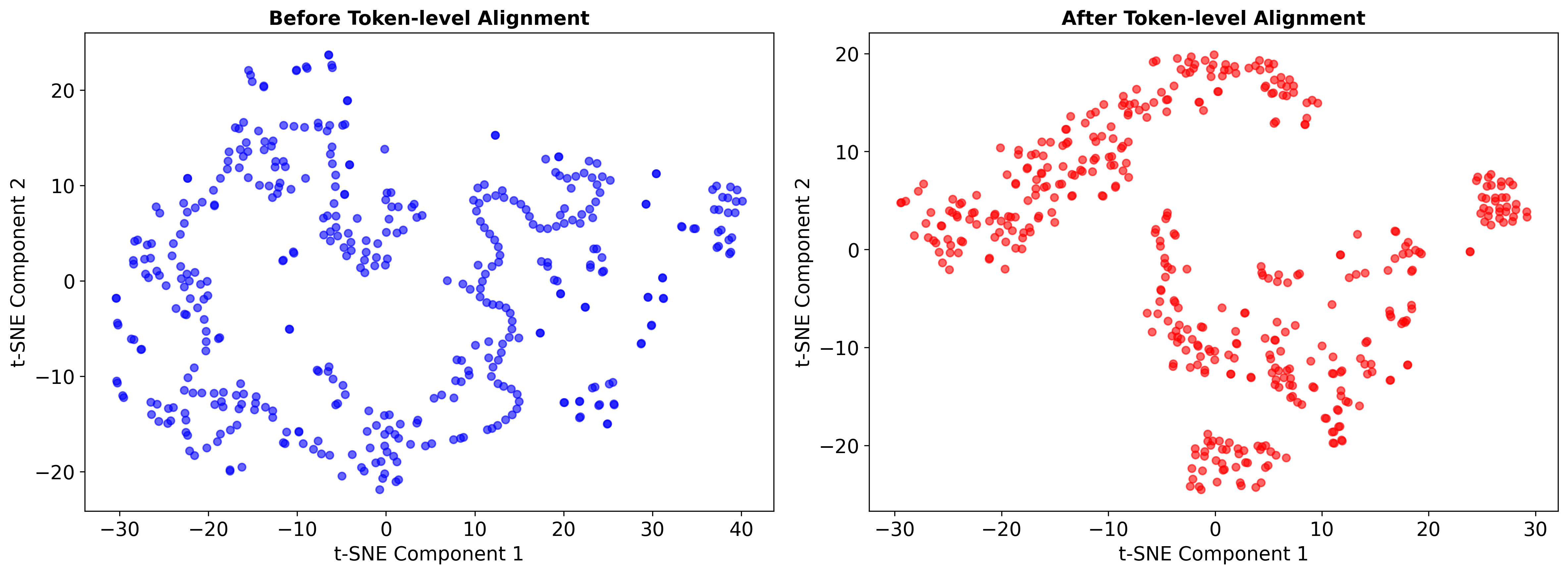}
  \caption{The t-SNE plots of time series embeddings before and after token-level alignment.}
\label{figure3}
\end{figure}

\textbf{t-SNE Visualization.} In order to evaluate the effectiveness of our proposed heterogeneous graph neural network, we also employ t-SNE visualization to examine the distributions of time series before and after token-level alignment. As shown in Figure \ref{figure3}, the raw time series embeddings before alignment exhibit dispersed and unstructured structure with noise and outliers in the latent space, indicating weak inherent patterns. After token-level alignment, we observe that the time series embeddings show well-separated, robust clusters, which reveal that the heterogeneous cross-modality module effectively endows time series with semantic information. 

\begin{figure}[htbp]
    \centering
    \begin{subfigure}[b]{0.23\textwidth} 
        \centering
        \includegraphics[width=\textwidth]{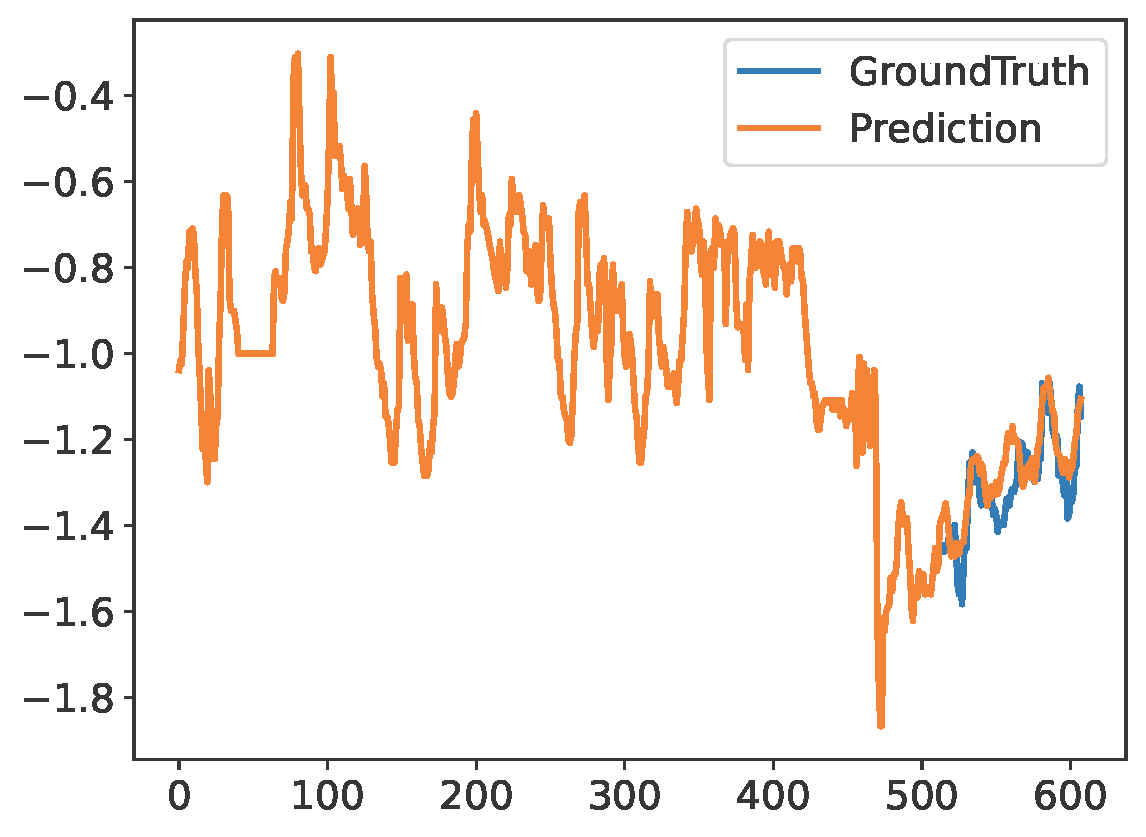}
        \caption*{(a) ETTh1}
        \label{fig:image1}
    \end{subfigure}
    \hfill
    \begin{subfigure}[b]{0.23\textwidth}   
        \centering
        \includegraphics[width=\textwidth]{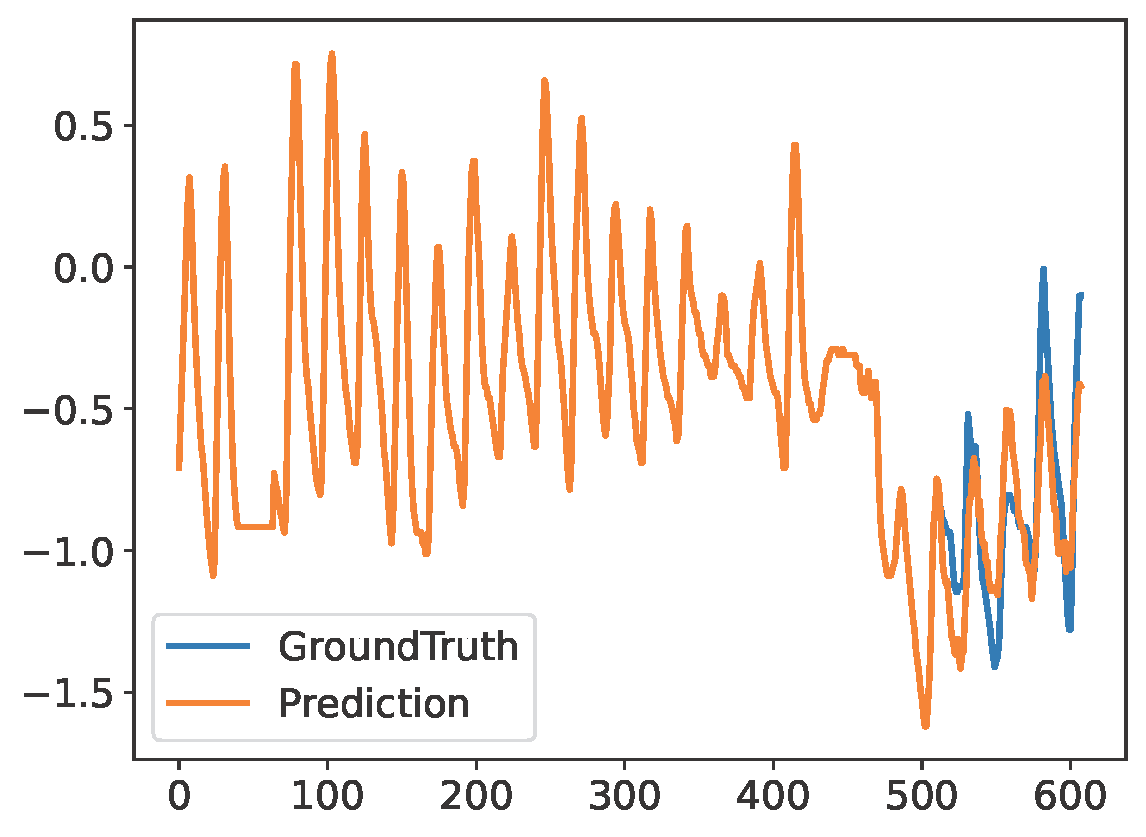}
        \caption*{(b) ETTh2}
        \label{fig:image2}
    \end{subfigure}
    \caption{Visualization of long-term forecasting performance.}
\label{figure4}
\end{figure}

\begin{figure}[htbp]
    \centering
        \begin{subfigure}[b]{0.23\textwidth} 
        \centering
        \includegraphics[width=\textwidth]{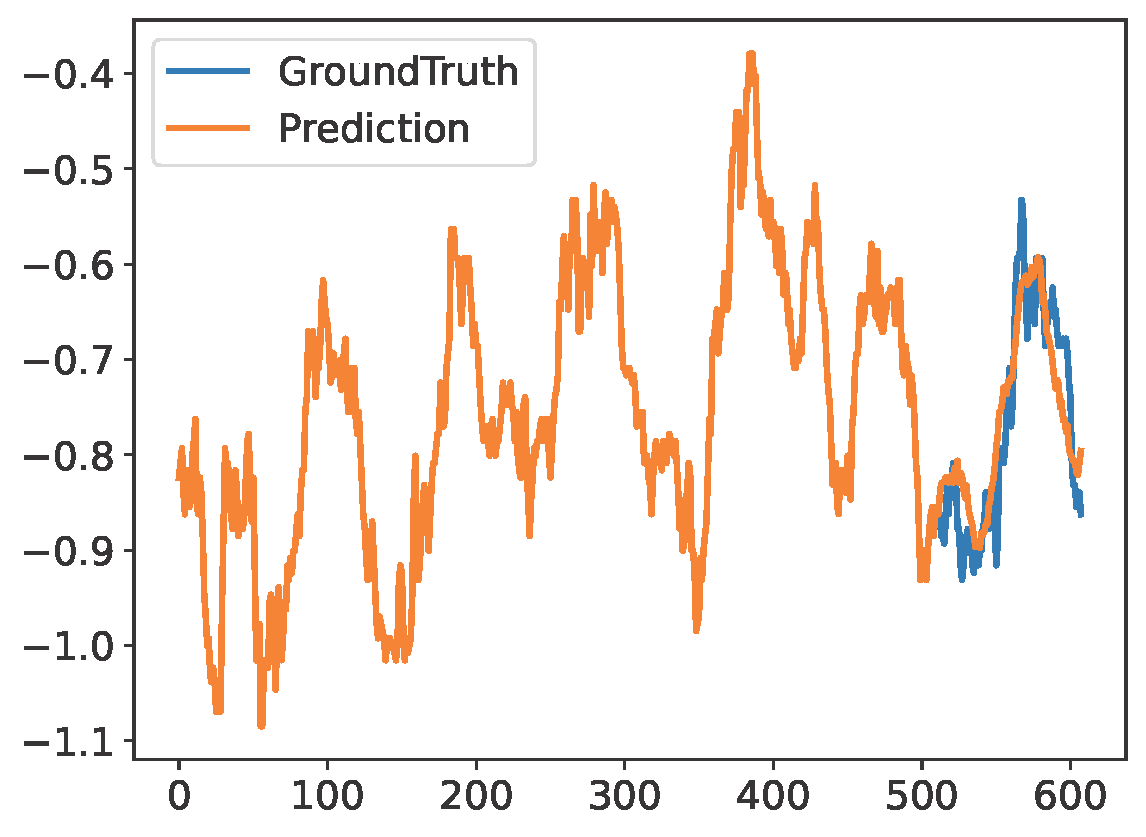}
        \caption*{(a) ETTm1}
        \label{fig:image5}
    \end{subfigure}
    \hfill
    \begin{subfigure}[b]{0.23\textwidth}  
        \centering
        \includegraphics[width=\textwidth]{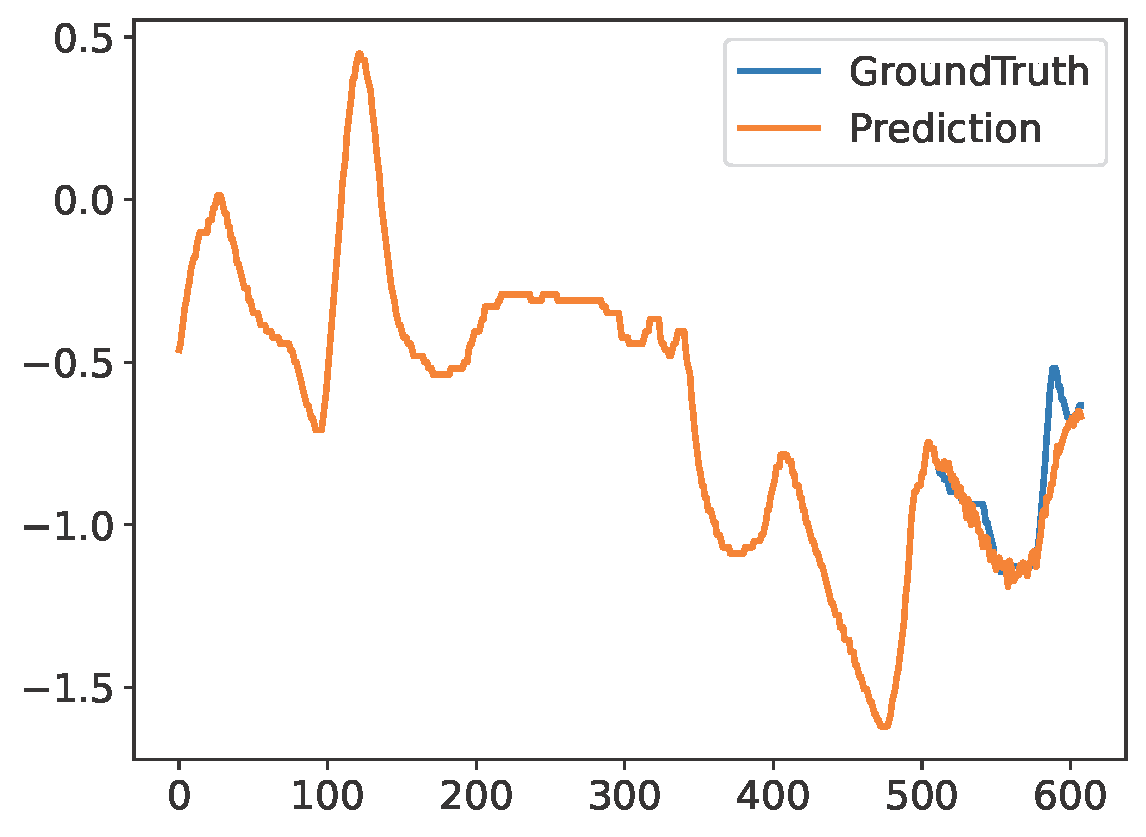}
        \caption*{(b) ETTm2}
        \label{fig:image8}
    \end{subfigure}
    \caption{Visualization of few-shot forecasting performance on 10\% training data.}
    \label{fig:all_images}
\label{figure5}
\end{figure}

\noindent \textbf{Visualization of Forecasting Results.} To explicitly observe the forecasting performance of the proposed model, we also visualize the long-term and few-shot forecasting results of the ETTh1 dataset with an input length of 512 and a prediction length of 96. As shown in Figure \ref{figure4} and Figure \ref{figure5}, FiCoTS achieves outstanding forecasting performance and precisely predicts potential fluctuations in the future.

\begin{figure}[htbp]
    \centering
    \begin{subfigure}[b]{0.23\textwidth} 
        \centering
        \includegraphics[width=\textwidth]{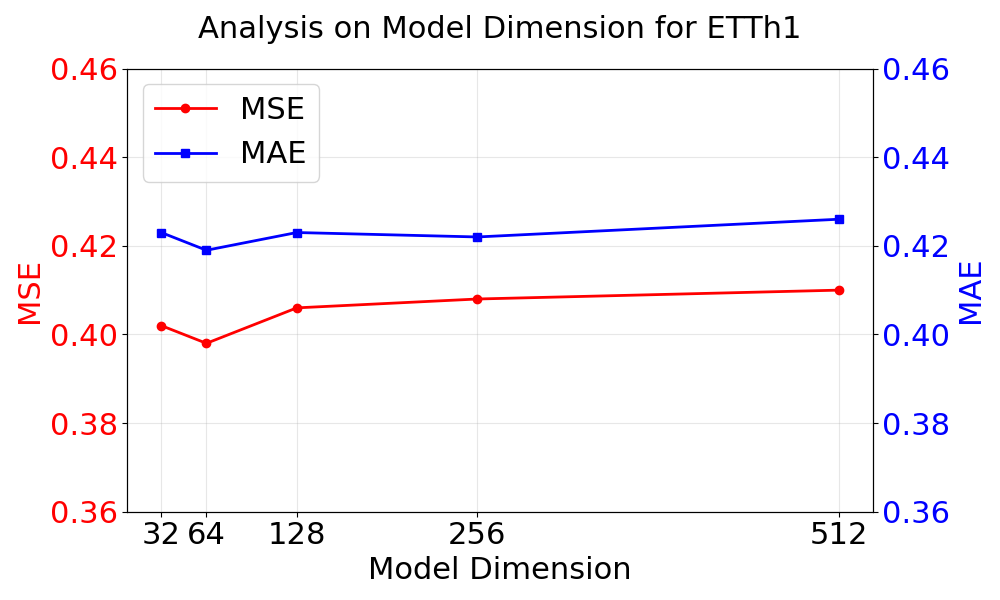}
        \caption*{\footnotesize{(a) Analysis on model dimension for ETTh1 dataset}}
        \label{fig:image1}
    \end{subfigure}
    \hfill
    \begin{subfigure}[b]{0.23\textwidth}   
        \centering
        \includegraphics[width=\textwidth]{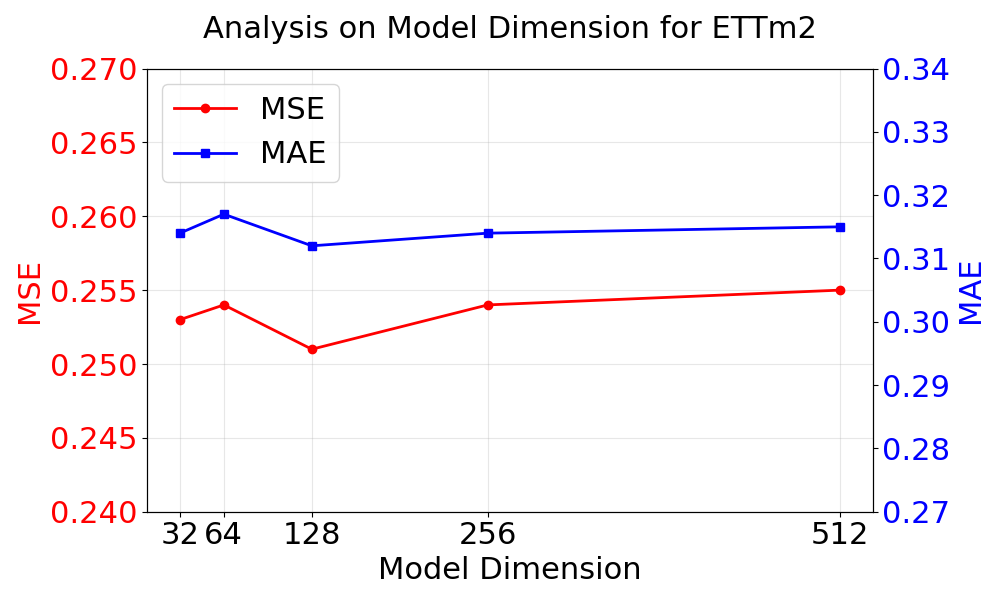}
        \caption*{\footnotesize{(b) Analysis on model dimension for ETTm2 dataset}}
        \label{fig:image2}
    \end{subfigure}
    
    \vspace{0.5cm}
    \begin{subfigure}[b]{0.23\textwidth} 
        \centering
        \includegraphics[width=\textwidth]{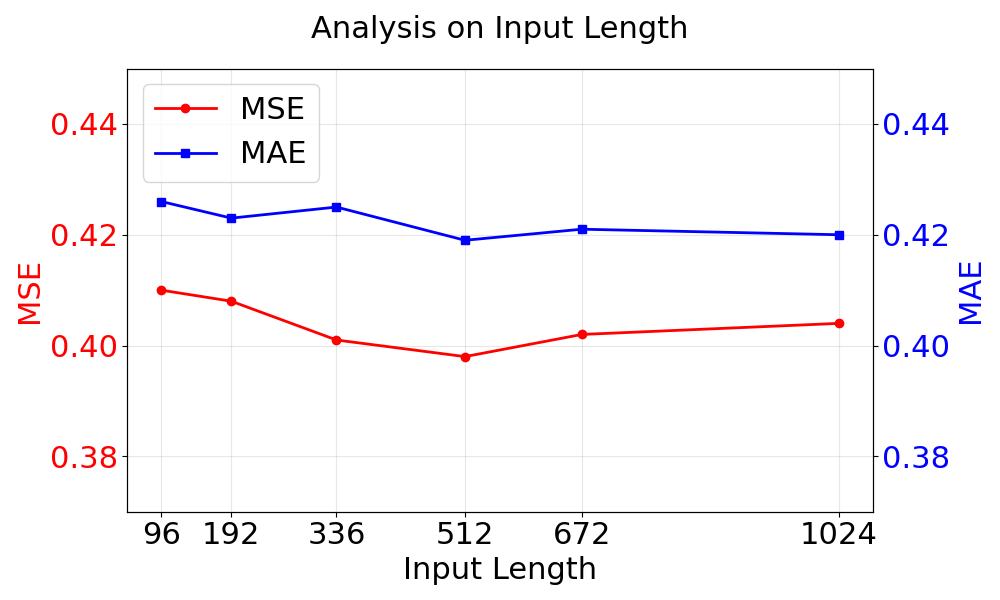}
        \caption*{\footnotesize{(c) Analysis on input length}}
        \label{fig:image3}
    \end{subfigure}
    \hfill
    \begin{subfigure}[b]{0.23\textwidth}   
        \centering
        \includegraphics[width=\textwidth]{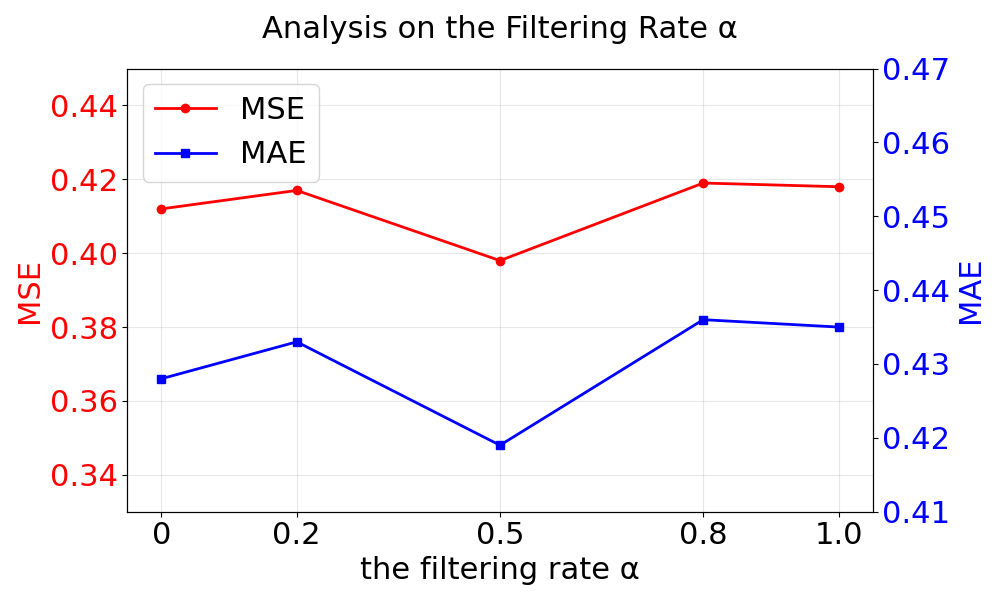}
        \caption*{\footnotesize{(d) Analysis on the filtering rate $\alpha$}}
        \label{fig:image4}
    \end{subfigure}    
    \caption{Hyperparameter sensitivity analysis.}
    \label{fig:all_images}\
\label{figure6}
\end{figure}

\noindent \textbf{Hyperparameter Sensitivity Analysis.} We conducted several experiments to examine the impact of main hyperparameters in FiCoTS: the dimension of the model, the input length, and the filtering sensitivity of the heterogeneous graph. As shown in Figure \ref{figure6} (a) and (b), the dimension of the model varies depending on the training datasets. For a smaller dataset like ETTh1, it is optimal to choose a dimension of 64; for a larger dataset like ETTm2, it is better to choose a larger dimension of 128. Figure \ref{figure6} (c) shows that the optimal input length is 512. Shorter and longer input strengths both result in performance degradation. This phenomenon is attributed to the fact that shorter input lacks sufficient temporal information, while longer input introduces excessive noise. Figure \ref{figure6} (d) presents the impact of the filtering sensitivity, which determines the structure of the heterogeneous graph and influences the effectiveness of modality alignment. We observed that 0.5 is the optimal sensitivity value for an appropriate filtering threshold. Smaller $\alpha$ results in extremely sparse links and contributes to inadequate modality interaction, while larger $\alpha$ may confuse the semantics of two modalities to some extent.

\section{Conclusion} 

This paper presents FiCoTS, a fine-to-coarse LLM-enhanced hierarchical cross-modality interaction framework for multimodal time series forecasting. Unlike recent LLM-as-Predictor models, we propose an LLM-as-Enhancer paradigm, which utilizes LLM to encode text modality and complement time series modality. With the three cross-modality interaction modules (i.e., token-level, feature-level, and decision-level) working together, FiCoTS facilitates progressive and comprehensive interaction between time series and text modalities. Extensive experiments on seven real-world benchmarks demonstrate that FiCoTS not only achieves superior state-of-the-art performance but also reduces computational costs. We believe that the fine-to-coarse cross-modality interaction framework provides an advanced paradigm for multimodal time series forecasting. 


\begin{acks}
To Robert, for the bagels and explaining CMYK and color spaces.
\end{acks}

\nocite{*} 

\clearpage

\appendix

\section{Experimental Details}

\subsection{Dataset Details}
\label{A1}
For long time forecasting task, we conduct experiments on seven datasets which are commonly used, including Electricity Transformer Temperature (ETT) datasets, Weather dataset, Electricity dataset and Traffic dataset. The dataset details are summarized in Table \ref{table5}. The ETT datasets record electricity transformer temperature sequences from two regions in China, further including four sub-datasets with two sampling rates, an hour for ETTh1, ETTh2 and 15 minutes for ETTm1 and ETTm2. Each dataset contains six power load variables and a targeted oil temperature variable. The Weather dataset is a collection of 21 weather indicators recorded every 10 minutes throughout 2020, including air temperature, humidity and so on. The Electricity dataset includes hourly electricity consumption data for 321 clients from 2016 to 2019. The Traffic dataset contains hourly road occupancy rates recorded by 862 sensors on the freeway across California from 2015 to 2016.

\begin{table}[h]
    \centering
    \small
    \tabcolsep=0.7mm
    \caption{Dataset Details.}
    \vspace{-1.5em}
    \resizebox{\linewidth}{!}{
    \begin{tabular}{l|ccccc}
        \shline
         Dataset Name & Variable & Series Length & Frequency &  Datasets Size & Domain \\
        \hline \hline
         ETTh1/ETTh2 & 7 & $\{96, 192, 336, 720\}$ & 1 hour & (34465, 11521, 11521) & Temperature \\
         ETTm1/ETTm2 & 7 & $\{96, 192, 336, 720\}$ & 15 min & (8545, 2881, 2881) & Temperature \\
         Weather & 21 & $\{96, 192, 336, 720\}$ & 1 hour & (36792, 5271, 10540) & Weather \\
         Electricity & 321 & $\{96, 192, 336, 720\}$ & 1 hour & (18317, 2633, 5261) & Electricity \\
         Traffic & 862 & $\{96, 192, 336, 720\}$ & 10 min & (12185, 1757, 3509) & Transportation \\
        \shline
    \end{tabular}
    }
    \label{tab:dataset}
    \vspace{-1.5em}
\label{table5}
\end{table}

\subsection{Baselines}

\noindent \textbf{TimesNet.} TimesNet converting 1-dimensional temporal feature to 2-dimensional space based on its multi-periodicity property. The proposed Timesblock captures inter- and intra-period temporal patterns in a residual way.\\
\noindent \textbf{Autoformer.} Autoformer replaces self-attention with an autocorrelation mechanism, efficiently capturing periodicity in time series.\\
\noindent \textbf{FEDformer.} FEDformer decomposes time series into trend and seasonal components. In addition, it utilizes frequency domain transformation to efficiently capture the global features of the time series.\\
\noindent \textbf{ETSformer.} ETSformer integrates exponential smoothing into the Transformer architecture to capture temporal trends and seasonal patterns, enhancing interpretability and forecasting performance.\\
\noindent \textbf{PatchTST.} PatchTST introduces patching strategy to segment time series into patches, leveraging Transformer layers to encode dependency between patches in a channel-independence way.\\
\noindent \textbf{iTransformer.} iTransformer inverts the traditional Transformer architecture and regards each variable as a patch. It leverages attention mechanism at the variable-level, capturing multivariate correlations.\\
\noindent \textbf{DLinear.} DLinear decomposes time series into trend and remainder components and processes them with separate linear layers, challenging the necessity of complex architectures.\\
\noindent \textbf{GPT4TS.} GPT4TS introduces the idea of leveraging LLM for time series analysis. It treats LLM as the backbone and designs linear tokenizer to adapt time series patches to the semantic space of LLM.\\
\noindent \textbf{TimeLLM.} Time-LLM utilizes LLM for time series analysis by a reprogramming strategy to transform time series patches into natural language tokens. It also adopts a Prompt-as-Prefix way to improve model performance.\\
\noindent \textbf{S$^{2}$IP-LLM.} S$^{2}$IP-LLM aligned the joint space of text and time series to fully activate LLM’s ability for time series tasks.The retrieved semantic anchors are regarded as prompts, enhancing the representation of time series.\\
\noindent \textbf{TimeVLM.} TimeVLM leverages pre-trained Vision-Language Models (VLMs) to unify temporal, visual, and textual modalities, enabling enhanced time series forecasting by transforming time series into images and designing textual prompts.\\

\subsection{Evaluation Metrics}
For long-term and few-shot forecasting task, we apply the Mean Squared Error(MSE) and Mean Absolute Error (MAE) as evaluation metrics. The  calculations of these evaluation metrics are as follows:
\begin{align*}
\mathrm{MSE} & =\frac{1}{H}\sum_{h=1}^{T}(\boldsymbol{Y}_{h}-\hat{\boldsymbol{Y}_{h}})^{2}, \quad\quad \ \ \mathrm{MAE}=\frac{1}{H}\sum_{h=1}^{H}|\boldsymbol{Y}_{h}-\hat{\boldsymbol{Y}}_{h}|.
\end{align*}

\section{Detailed Textual Prompt}
\label{B}
\subsection{Dataset Description.} Below are the descriptions of the seven datasets we used:\\
\noindent \textbf{ETT.} The Electricity Transformer Temperature (ETT) is a crucial indicator in the electric power long-term deployment. This dataset consists of 2 years data from two separated counties in China. To explore the granularity on the long sequence time series forecasting problem, different subsets are created, {ETTh1, ETTh2} for 1-hour-level and ETTm1 for 15-minutes-level. Each data point consists of the target value oil temperature and 6 power load features. The train/val/test is 12/4/4 months.\\
\noindent \textbf{Weather.} Weather is recorded every 10 minutes for the 2020 whole year, which contains 21 meteorological indicators, such as air temperature, humidity, etc.\\
\noindent \textbf{Electricity.} Measurements of electric power consumption in one household with a one-minute sampling rate over a period of almost 4 years. Different electrical quantities and some sub-metering values are available.This archive contains 2075259 measurements gathered in a house located in Sceaux (7km of Paris, France) between December 2006 and November 2010 (47 months).\\
\noindent \textbf{Traffic.} Traffic is a collection of hourly data from California Department of Transportation, which describes the road occupancy rates measured by different sensors on San Francisco Bay area freeways.\\

\subsection{Task Instruction.} Forecast the next $\{pred\_len\}$ steps using the past $\{seq\_len\}$ steps.

\subsection{Input Statistics.} Input statistics: min value = $\{min\_value\}$, max value = $\{max\_value\}$, median value = $\{median\_value\}$, the overall trend is $\{trend\_direction\}$.

\begin{table*}[t]
    \centering
    \small
    \tabcolsep=0.8mm
    \caption{Full long-term forecasting results. The input sequence length is set to 512. The predictive lengths are set to $\{96, 192, 336, 720\}$. We bold the best performance and underline the second best performance.}
    \vspace{-1.5em}
    \resizebox{\textwidth}{!}{
    \begin{tabular}{cc|cc|cc|cc|cc|cc||cc|cc||cc|cc|cc|cc|cc}
    \shline
    \multicolumn{2}{c}{\multirow{3}{*}{Method}} & \multicolumn{10}{c||}{\textbf{\textit{LLM-based}}} & \multicolumn{4}{c||}{\textbf{\textit{Non-Transformer-based}}} & \multicolumn{10}{c}{\textbf{\textit{Transformer-based}}} \\ \cline{3-26}
    \multicolumn{2}{c}{} & \multicolumn{2}{c|}{FiCoTS} & \multicolumn{2}{c|}{TimeVLM} & \multicolumn{2}{c|}{S$^{2}$IP-LLM} & \multicolumn{2}{c|}{TimeLLM} & \multicolumn{2}{c||}{GPT4TS} & \multicolumn{2}{c|}{TimesNet} & \multicolumn{2}{c||}{DLinear} & \multicolumn{2}{c|}{PatchTST} & \multicolumn{2}{c|}{iTransformer} & \multicolumn{2}{c|}{FEDformer} & \multicolumn{2}{c|}{Autoformer} & \multicolumn{2}{c}{ETSformer} \\ \cline{3-26} 
    \multicolumn{2}{c}{} & MSE & MAE & MSE & MAE & MSE & MAE & MSE & MAE & MSE & MAE & MSE & MAE & MSE & MAE & MSE & MAE & MSE & MAE & MSE & MAE & MSE & MAE & MSE & MAE\\
    \hline \hline
    \multicolumn{1}{c|}{\multirow{5}{*}{\rotatebox{90}{ETTh1}}} & 96 & \underline{0.363} & \underline{0.392} & \textbf{0.361} & \textbf{0.386} & 0.367 & 0.398 & 0.383 & 0.404 & 0.376 & 0.397 & 0.384 & 0.402 & 0.375 & 0.399 & 0.370 & 0.399 & 0.395 & 0.420 & 0.376 & 0.419 & 0.449 & 0.459 & 0.494 & 0.479 \\
    \multicolumn{1}{c|}{} & 192 & \textbf{0.396} & \textbf{0.412} & \underline{0.397} & \underline{0.415} & 0.402 & 0.422 & 0.427 & 0.431 & 0.416 & 0.418 & 0.436 & 0.429 & 0.405 & 0.416 & 0.413 & 0.421 & 0.427 & 0.441 & 0.420 & 0.448 & 0.500 & 0.482 & 0.538 & 0.504 \\
    \multicolumn{1}{c|}{} & 336 & \textbf{0.413} & \underline{0.425} & \underline{0.420} & \textbf{0.421} & 0.432 & 0.451 & 0.430 & 0.436 & 0.442 & 0.433 & 0.491 & 0.469 & 0.439 & 0.443 & 0.422 & 0.436 & 0.445 & 0.457 & 0.459 & 0.465 & 0.521 & 0.496 & 0.574 & 0.521\\
    \multicolumn{1}{c|}{} & 720 & \textbf{0.420} & \textbf{0.447} & \underline{0.441} & \underline{0.458} & 0.472 & 0.474 & 0.465 & 0.469 & 0.477 & 0.456 & 0.521 & 0.500 & 0.472 & 0.490 & 0.447 & 0.466 & 0.537 & 0.530 & 0.506 & 0.507 & 0.514 & 0.512 & 0.562 & 0.535 \\ \cline{2-26}
    \multicolumn{1}{c|}{} & Avg & \textbf{0.398} & \textbf{0.419} & \underline{0.405} & \underline{0.420} & 0.418 & 0.436 & 0.426 & 0.435 & 0.427 & 0.436 & 0.458 & 0.450 & 0.422 & 0.437 & 0.413 & 0.430 & 0.451 & 0.462 & 0.440 & 0.460 & 0.496 & 0.487 & 0.542 & 0.510\\
    \hline \hline
    \multicolumn{1}{c|}{\multirow{5}{*}{\rotatebox{90}{ETTh2}}} & 96 & \textbf{0.267} & \textbf{0.334} & \textbf{0.267} & \underline{0.335} & 0.284 & 0.345 & 0.293 & 0.348 & 0.285 & 0.342 & 0.340 & 0.374 & 0.289 & 0.353 & \underline{0.274} & 0.336 & 0.304 & 0.360 & 0.358 & 0.397 & 0.346 & 0.388 & 0.340 & 0.391\\
    \multicolumn{1}{c|}{} & 192 & \textbf{0.326} & \underline{0.374} & \textbf{0.326} & \textbf{0.373} & 0.349 & 0.387 & 0.356 & 0.391 & 0.354 & 0.389 & 0.402 & 0.414 & 0.383 & 0.418 & \underline{0.339} & 0.379 & 0.377 & 0.403 & 0.429 & 0.439 & 0.456 & 0.452 & 0.430 & 0.439\\
    \multicolumn{1}{c|}{} & 336 & \underline{0.353} & \underline{0.401} & 0.357 & 0.406 & 0.368 & 0.417 & 0.372 & 0.408 & 0.373 & 0.407 & 0.452 & 0.452 & 0.448 & 0.465 & \textbf{0.329} & \textbf{0.380} & 0.405 & 0.429 & 0.496 & 0.487 & 0.482 & 0.486 & 0.485 & 0.479 \\
    \multicolumn{1}{c|}{} & 720 & \underline{0.388} & \underline{0.432} & 0.412 & 0.449 & 0.419 & 0.445 & 0.421 & 0.446 & 0.406 & 0.441 & 0.462 & 0.468 & 0.605 & 0.551 & \textbf{0.379} & \textbf{0.422} & 0.443 & 0.464 & 0.463 & 0.474 & 0.450 & 0.459 & 0.439 & 0.452\\ \cline{2-26}
    \multicolumn{1}{c|}{} & Avg & \underline{0.335} & \underline{0.385} & 0.341 & 0.391 & 0.355 & 0.399 & 0.361 & 0.398 & 0.354 & 0.394 & 0.414 & 0.427 & 0.431 & 0.446 & \textbf{0.330} & \textbf{0.379} & 0.382 & 0.414 & 0.437 & 0.449 & 0.450 & 0.459 & 0.439 & 0.452 \\
    \hline \hline
    \multicolumn{1}{c|}{\multirow{5}{*}{\rotatebox{90}{ETTm1}}} & 96 & \textbf{0.290} & \textbf{0.340} & 0.304 & 0.346 & \underline{0.291} & 0.348 & 0.294 & 0.345 & 0.292 & 0.346 & 0.338 & 0.375 & 0.299 & 0.343 & \textbf{0.290} & \underline{0.342} & 0.312 & 0.366 & 0.379 & 0.419 & 0.505 & 0.475 & 0.375 & 0.398 \\
    \multicolumn{1}{c|}{} & 192 & \textbf{0.325} & \textbf{0.362} & 0.332 & \textbf{0.366} & \textbf{0.323} & 0.368 & 0.330 & 0.368 & 0.332 & 0.372 & 0.374 & 0.387 & 0.335 & 0.365 & 0.332 & 0.369 & 0.347 & 0.385 & 0.426 & 0.441 & 0.553 & 0.496 & 0.408 & 0.410 \\
    \multicolumn{1}{c|}{} & 336 & \textbf{0.360} & \textbf{0.380} & 0.364 & \textbf{0.383} & \textbf{0.361} & 0.392 & 0.365 & 0.392 & 0.366 & 0.394 & 0.410 & 0.411 & 0.369 & 0.386 & 0.366 & 0.392 & 0.379 & 0.404 & 0.445 & 0.459 & 0.621 & 0.537 & 0.435 & 0.428\\
    \multicolumn{1}{c|}{} & 720 & \textbf{0.410} & \textbf{0.411} & \textbf{0.402} & \textbf{0.410} & \textbf{0.410} & 0.420 & 0.427 & 0.431 & 0.417 & 0.427 & 0.478 & 0.450 & 0.425 & 0.421 & 0.416 & 0.420 & 0.441 & 0.442 & 0.543 & 0.490 & 0.671 & 0.561 & 0.499 & 0.462 \\ \cline{2-26}
    \multicolumn{1}{c|}{} & Avg & \textbf{0.346} & \textbf{0.373} & \underline{0.350} & \underline{0.377} & 0.346 & 0.382 & 0.354 & 0.384 & 0.352 & 0.383 & 0.400 & 0.406 & 0.357 & 0.378 & 0.351 & 0.380 & 0.370 & 0.399 & 0.448 & 0.452 & 0.588 & 0.517 & 0.429 & 0.425 \\
    \hline \hline
    \multicolumn{1}{c|}{\multirow{5}{*}{\rotatebox{90}{ETTm2}}} & 96 & \underline{0.164} & \underline{0.255} & \textbf{0.160} & \textbf{0.250} & 0.167 & 0.257 & 0.175 & 0.265 & 0.173 & 0.262 & 0.187 & 0.267 & 0.167 & 0.269 & 0.165 & 0.255 & 0.179 & 0.271 & 0.203 & 0.287 & 0.255 & 0.339 & 0.189 & 0.280 \\
    \multicolumn{1}{c|}{} & 192 & \underline{0.217} & \textbf{0.291} & \textbf{0.215} & \textbf{0.291} & 0.227 & 0.303 & 0.243 & 0.316 & 0.229 & 0.301 & 0.249 & 0.309 & 0.224 & 0.303 & 0.220 & \underline{0.292} & 0.242 & 0.313 & 0.269 & 0.328 & 0.281 & 0.340 & 0.253 & 0.319 \\
    \multicolumn{1}{c|}{} & 336 & \textbf{0.270} & \textbf{0.325} & \textbf{0.270} & \textbf{0.325} & 0.285 & 0.346 & 0.294 & 0.343 & 0.286 & 0.341 & 0.321 & 0.351 & 0.281 & 0.342 & 0.274 & 0.329 & 0.288 & 0.344 & 0.325 & 0.366 & 0.339 & 0.372 & 0.314 & 0.357 \\
    \multicolumn{1}{c|}{} & 720 & \underline{0.354} & \textbf{0.378} & \textbf{0.348} & \textbf{0.378} & 0.368 & 0.398 & 0.389 & 0.410 & 0.378 & 0.401 & 0.408 & 0.403 & 0.397 & 0.421 & 0.362 & \underline{0.385} & 0.378 & 0.397 & 0.421 & 0.415 & 0.433 & 0.432 & 0.414 & 0.413 \\ \cline{2-26}
    \multicolumn{1}{c|}{} & Avg & \underline{0.251} & \underline{0.312} & \textbf{0.248} & \textbf{0.311} & 0.262 & 0.326 & 0.275 & 0.334 & 0.266 & 0.326 & 0.291 & 0.333 & 0.267 & 0.333 & 0.255 & 0.315 & 0.272 & 0.331 & 0.305 & 0.349 & 0.327 & 0.371 & 0.293 & 0.342 \\
    \hline \hline
    \multicolumn{1}{c|}{\multirow{5}{*}{\rotatebox{90}{Weather}}} & 96 & \textbf{0.148} & \textbf{0.197} & \textbf{0.148} & \underline{0.200} & \underline{0.149} & \underline{0.200} & 0.163 & 0.210 & 0.162 & 0.212 & 0.172 & 0.220 & 0.176 & 0.237 & 0.149 & 0.198 & 0.253 & 0.304 & 0.217 & 0.296 & 0.276 & 0.336 & 0.197 & 0.281 \\
    \multicolumn{1}{c|}{} & 192 & \textbf{0.192} & \textbf{0.239} & \underline{0.193} & \underline{0.240} & 0.195 & 0.244 & 0.205 & 0.245 & 0.204 & 0.248 & 0.219 & 0.261 & 0.220 & 0.282 & 0.194 & 0.241 & 0.280 & 0.319 & 0.276 & 0.336 & 0.307 & 0.367 & 0.237 & 0.312 \\
    \multicolumn{1}{c|}{} & 336 & \underline{0.244} & \underline{0.281} & \textbf{0.243} & \underline{0.281} & 0.246 & \textbf{0.280} & 0.257 & 0.287 & 0.254 & 0.286 & 0.280 & 0.306 & 0.265 & 0.319 & 0.245 & 0.282 & 0.321 & 0.344 & 0.339 & 0.380 & 0.359 & 0.395 & 0.298 & 0.353 \\
    \multicolumn{1}{c|}{} & 720 & \underline{0.315} & \textbf{0.331} & \textbf{0.312} & \underline{0.332} & 0.320 & 0.336 & 0.323 & 0.332 & 0.326 & 0.337 & 0.365 & 0.359 & 0.333 & 0.362 & 0.314 & 0.334 & 0.364 & 0.374 & 0.403 & 0.428 & 0.419 & 0.428 & 0.352 & 0.288 \\ \cline{2-26}
    \multicolumn{1}{c|}{} & Avg & \textbf{0.224} & \textbf{0.262} & \textbf{0.224} & \underline{0.263} & 0.228 & 0.265 & 0.237 & 0.269 & 0.237 & 0.270 & 0.259 & 0.287 & 0.248 & 0.300 & 0.225 & 0.264 & 0.304 & 0.335 & 0.309 & 0.360 & 0.338 & 0.382 & 0.271 & 0.334 \\
    \hline \hline
    \multicolumn{1}{c|}{\multirow{5}{*}{\rotatebox{90}{Electricity}}} & 96 & \underline{0.132} & \underline{0.227} & 0.142 & 0.245 & 0.138 & 0.234 & 0.140 & 0.236 & 0.139 & 0.238 & 0.168 & 0.272 & 0.140 & 0.237 & \textbf{0.129} & \textbf{0.222} & 0.147 & 0.248 & 0.193 & 0.308 & 0.201 & 0.317 & 0.187 & 0.304 \\
    \multicolumn{1}{c|}{} & 192 & \textbf{0.148} & \underline{0.241} & 0.157 & 0.260 & 0.153 & 0.252 & \underline{0.150} & 0.249 & 0.153 & 0.251 & 0.184 & 0.289 & 0.153 & 0.249 & 0.157 & \textbf{0.240} & 0.165 & 0.267 & 0.201 & 0.315 & 0.222 & 0.334 & 0.199 & 0.315\\
    \multicolumn{1}{c|}{} & 336 & \underline{0.164} & \textbf{0.258} & 0.174 & 0.276 & 0.169 & 0.270 & 0.168 & 0.267 & 0.169 & 0.266 & 0.198 & 0.300 & 0.169 & 0.267 & \textbf{0.163} & \underline{0.259} & 0.178 & 0.279 & 0.214 & 0.329 & 0.231 & 0.338 & 0.212 & 0.329\\
    \multicolumn{1}{c|}{} & 720 & \underline{0.203} & \underline{0.291} & 0.214 & 0.308 & 0.204 & 0.293 & 0.209 & 0.302 & 0.206 & 0.297 & 0.220 & 0.320 & \underline{0.203} & 0.301 & \textbf{0.197} & \textbf{0.290} & 0.322 & 0.398 & 0.246 & 0.355 & 0.254 & 0.361 & 0.233 & 0.345\\ \cline{2-26}
    \multicolumn{1}{c|}{} & Avg & \textbf{0.161} & \underline{0.254} & 0.172 & 0.273 & 0.166 & 0.262 & 0.167 & 0.264 & 0.167 & 0.263 & 0.192 & 0.295 & 0.166 & 0.263 & \textbf{0.161} & \textbf{0.252} & 0.203 & 0.298 & 0.214 & 0.327 & 0.227 & 0.338 & 0.208 & 0.323 \\
    \hline \hline
    \multicolumn{1}{c|}{\multirow{5}{*}{\rotatebox{90}{Traffic}}} & 96 & 0.372 & \underline{0.259} & 0.393 & 0.290 & 0.379 & 0.274 & 0.384 & 0.278 & 0.388 & 0.282 & 0.593 & 0.321 & 0.410 & 0.282 & \textbf{0.360} & \textbf{0.249} & \underline{0.367} & 0.288 & 0.587 & 0.366 & 0.613 & 0.388 & 0.607 & 0.392 \\
    \multicolumn{1}{c|}{} & 192 & 0.398 & \underline{0.270} & 0.405 & 0.296 & 0.397 & 0.282 & 0.398 & 0.286 & 0.407 & 0.290 & 0.617 & 0.336 & 0.423 & 0.287 & \underline{0.379} & \textbf{0.256} & \textbf{0.378} & 0.293 & 0.604 & 0.373 & 0.616 & 0.382 & 0.621 & 0.399 \\
    \multicolumn{1}{c|}{} & 336 & 0.405 & \underline{0.275} & 0.420 & 0.305 & 0.407 & 0.289 & 0.408 & 0.289 & 0.412 & 0.294 & 0.629 & 0.336 & 0.436 & 0.296 & \underline{0.392} & \textbf{0.264} & \textbf{0.389} & 0.294 & 0.621 & 0.383 & 0.622 & 0.337 & 0.622 & 0.396 \\
    \multicolumn{1}{c|}{} & 720 & 0.437 & \underline{0.292} & 0.459 & 0.323 & 0.440 & 0.301 & 0.436 & 0.303 & 0.450 & 0.312 & 0.640 & 0.350 & 0.466 & 0.315 & \underline{0.432} & \textbf{0.286} & \textbf{0.401} & 0.304 & 0.626 & 0.382 & 0.660 & 0.408 & 0.632 & 0.396 \\ \cline{2-26}
    \multicolumn{1}{c|}{} & Avg & 0.403 & \underline{0.274} & 0.419 & 0303 & 0.405 & 0.286 & 0.407 & 0.289 & 0.414 & 0.294 & 0.620 & 0.336 & 0.433 & 0.295 & \underline{0.390} & \textbf{0.263} & \textbf{0.389} & 0.295 & 0.610 & 0.376 & 0.628 & 0.379 & 0.621 & 0.396 \\
    \shline
    
    \end{tabular}
    }
\label{table6}
\end{table*}

\begin{table*}[h]
    \centering
    \small
    \tabcolsep=0.8mm
    \caption{Full results of few-shot learning on 10\% training data. The input sequence length is set to 512. The predictive lengths are set to $\{96, 192, 336, 720\}$. We bold the best performance and underline the second best performance.}
   \vspace{-1.5em}
    \resizebox{\textwidth}{!}{
    \begin{tabular}{cc|cc|cc|cc|cc|cc||cc|cc||cc|cc|cc|cc|cc}
    \shline
    \multicolumn{2}{c}{\multirow{3}{*}{Method}} & \multicolumn{10}{c||}{\textbf{\textit{LLM-based}}} & \multicolumn{4}{c||}{\textbf{\textit{Non-Transformer-based}}} & \multicolumn{10}{c}{\textbf{\textit{Transformer-based}}} \\ \cline{3-26}
    \multicolumn{2}{c}{} & \multicolumn{2}{c|}{FiCoTS} & \multicolumn{2}{c|}{TimeVLM} & \multicolumn{2}{c|}{S$^{2}$IP-LLM} & \multicolumn{2}{c|}{TimeLLM} & \multicolumn{2}{c||}{GPT4TS} & \multicolumn{2}{c|}{TimesNet} & \multicolumn{2}{c||}{DLinear} & \multicolumn{2}{c|}{PatchTST} & \multicolumn{2}{c|}{iTransformer} & \multicolumn{2}{c|}{FEDformer} & \multicolumn{2}{c|}{Autoformer} & \multicolumn{2}{c}{ETSformer} \\ \cline{3-26}
    \multicolumn{2}{c}{} & MSE & MAE & MSE & MAE & MSE & MAE & MSE & MAE & MSE & MAE & MSE & MAE & MSE & MAE & MSE & MAE & MSE & MAE & MSE & MAE & MSE & MAE & MSE & MAE\\
    \hline \hline
    \multicolumn{1}{c|}{\multirow{5}{*}{\rotatebox{90}{ETTh1}}} & 96  & \textbf{0.385} & \underline{0.410} & \underline{0.391} & \textbf{0.404} & 0.481 & 0.474 & 0.720 & 0.533 & 0.458 & 0.456 & 0.861 & 0.628 & 0.492 & 0.495 & 0.516 & 0.485 & 0.790 & 0.586 & 0.512 & 0.499 & 0.613 & 0.552  & 1.112 & 0.806\\
    \multicolumn{1}{c|}{} & 192 & \textbf{0.411} & \textbf{0.427} & \underline{0.420} & \underline{0.431} & 0.518 & 0.491 & 0.747 & 0.545 & 0.570 & 0.516 & 0.797 & 0.593 & 0.565 & 0.538 & 0.598 & 0.524 & 0.837 & 0.609 & 0.624 & 0.555 & 0.722 & 0.598  & 1.155 & 0.823\\
    \multicolumn{1}{c|}{} & 336 & \textbf{0.429} & \textbf{0.442} & \underline{0.439} & \underline{0.448} & 0.664 & 0.570 & 0.793 & 0.551 & 0.608 & 0.535 & 0.941 & 0.648 & 0.721 & 0.622 & 0.657 & 0.550 & 0.780 & 0.575 & 0.691 & 0.574 & 0.750 & 0.619  & 1.179 & 0.832\\
    \multicolumn{1}{c|}{} & 720 & \textbf{0.460} & \textbf{0.476} & \underline{0.476} & \underline{0.484} & 0.711 & 0.584 & 0.880 & 0.584 & 0.725 & 0.591 & 0.877 & 0.641 & 0.986 & 0.743 & 0.762 & 0.610 & 1.234 & 0.811 & 0.728 & 0.614 & 0.721 & 0.616  & 1.273 &
    0.874\\ \cline{2-26}
    \multicolumn{1}{c|}{} & Avg & \textbf{0.421} & \textbf{0.438} & \underline{0.431} & \underline{0.442} & 0.593 & 0.529 & 0.785 & 0.553 & 0.590 & 0.525 & 0.869 & 0.628 & 0.691 & 0.600 & 0.633 & 0.542 & 0.910 & 0.860 & 0.639 & 0.561 & 0.702 & 0.596  & 1.180 & 0.834\\
    \hline \hline
    \multicolumn{1}{c|}{\multirow{5}{*}{\rotatebox{90}{ETTh2}}} & 96  & \textbf{0.277} & \textbf{0.342} & \underline{0.284} & \underline{0.347} & 0.354 & 0.400 & 0.334 & 0.381 & 0.331 & 0.374 & 0.378 & 0.409 & 0.357 & 0.411 & 0.353 & 0.389 & 0.404 & 0.435 & 0.382 & 0.416 & 0.413 & 0.451  & 0.678 & 0.619\\
    \multicolumn{1}{c|}{} & 192 & \textbf{0.334} & \textbf{0.378} & \underline{0.349} & \underline{0.398} & 0.401 & 0.423 & 0.430 & 0.438 & 0.402 & 0.411 & 0.490 & 0.467 & 0.569 & 0.519 & 0.403 & 0.414 & 0.470 & 0.474 & 0.478 & 0.474 & 0.474 & 0.477  & 0.785 & 0.666\\
    \multicolumn{1}{c|}{} & 336 & \textbf{0.361} & \textbf{0.404} & \underline{0.370} & \underline{0.412} & 0.442 & 0.450 & 0.449 & 0.458 & 0.406 & 0.537 & 0.494 & 0.433 & 0.671 & 0.572 & 0.426 & 0.441 & 0.489 & 0.485 & 0.504 & 0.501 & 0.547 & 0.543 & 0.839 & 0.694\\
    \multicolumn{1}{c|}{} & 720 & \textbf{0.394} & \textbf{0.433} & \underline{0.441} & \underline{0.466} & 0.480 & 0.486 & 0.485 & 0.490 & 0.449 & 0.464 & 0.510 & 0.491 & 0.824 & 0.648 & 0.477 & 0.480 & 0.593 & 0.538 & 0.499 & 0.509 & 0.516 & 0.523 & 1.273 &
    0.874\\ \cline{2-26}
    \multicolumn{1}{c|}{} & Avg & \textbf{0.341} & \textbf{0.389} & \underline{0.361} & \underline{0.405} & 0.419 & 0.439 & 0.424 & 0.441 & 0.397 & 0.421 & 0.479 & 0.465 & 0.605 & 0.538 & 0.415 & 0.431 & 0.489 & 0.483 & 0.466 & 0.475 & 0.488 & 0.499 & 0.894 & 0.713\\
    \hline \hline
    \multicolumn{1}{c|}{\multirow{5}{*}{\rotatebox{90}{ETTm1}}} & 96  & \textbf{0.309} & \textbf{0.353} & \underline{0.310} & \underline{0.354} & 0.388 & 0.401 & 0.412 & 0.422 & 0.390 & 0.404 & 0.583 & 0.501 & 0.352 & 0.392 & 0.410 & 0.419 & 0.709 & 0.556 & 0.578 & 0.518 & 0.774 & 0.614 & 0.911 & 0.688\\
    \multicolumn{1}{c|}{} & 192 & \textbf{0.339} & \textbf{0.369} & \underline{0.340} & \underline{0.370} & 0.422 & 0.421 & 0.447 & 0.438 & 0.429 & 0.423 & 0.630 & 0.528 & 0.382 & 0.412 & 0.437 & 0.434 & 0.717 & 0.548 & 0.617 & 0.546 & 0.754 & 0.592 & 0.955 & 0.703\\
    \multicolumn{1}{c|}{} & 336 & \underline{0.370} & \underline{0.389} & \textbf{0.369} & \textbf{0.387} & 0.456 & 0.430 & 0.497 & 0.465 & 0.469 & 0.439 & 0.725 & 0.568 & 0.419 & 0.434 & 0.476 & 0.454 & 0.735 & 0.575 & 0.998 & 0.775 & 0.869 & 0.677 & 0.991 & 0.719\\
    \multicolumn{1}{c|}{} & 720 & \underline{0.424} & \underline{0.419} & \textbf{0.423} & \textbf{0.417} & 0.554 & 0.490 & 0.594 & 0.521 & 0.569 & 0.498 & 0.769 & 0.549 & 0.490 & 0.477 & 0.681 & 0.556 & 0.752 & 0.584 & 0.693 & 0.579 & 0.810 & 0.630 & 1.062 & 0.747\\ \cline{2-26}
    \multicolumn{1}{c|}{} & Avg & \textbf{0.360} & \textbf{0.382} & \textbf{0.360} & \textbf{0.382} & 0.455 & 0.435 & 0.487 & 0.461 & 0.464 & 0.441 & 0.677 & 0.537 & \underline{0.411} & \underline{0.429} & 0.501 & 0.466 & 0.728 & 0.565 & 0.722 & 0.605 & 0.802 & 0.628 & 0.980 & 0.714\\
    \hline \hline
    \multicolumn{1}{c|}{\multirow{5}{*}{\rotatebox{90}{ETTm2}}} & 96  & \textbf{0.168} & \textbf{0.258} & \underline{0.169} & \underline{0.260} & 0.192 & 0.274 & 0.224 & 0.296 & 0.188 & 0.269 & 0.212 & 0.285 & 0.213 & 0.303 & 0.191 & 0.274 & 0.245 & 0.322 & 0.291 & 0.399 & 0.352 & 0.454 & 0.331 & 0.430\\
    \multicolumn{1}{c|}{} & 192 & \textbf{0.222} & \textbf{0.296} & \textbf{0.222} & \textbf{0.296} & \underline{0.246} & 0.313 & 0.260 & 0.317 & 0.251 & \underline{0.309} & 0.270 & 0.323 & 0.278 & 0.345 & 0.252 & 0.317 & 0.274 & 0.338 & 0.307 & 0.379 & 0.694 & 0.691 & 0.400 & 0.464\\
    \multicolumn{1}{c|}{} & 336 & \textbf{0.275} & \textbf{0.329} & \underline{0.278} & \underline{0.335} & 0.301 & 0.340 & 0.312 & 0.349 & 0.307 & 0.346 & 0.323 & 0.353 & 0.338 & 0.385 & 0.306 & 0.353 & 0.361 & 0.394 & 0.543 & 0.559 & 2.408 & 1.407 & 0.469 & 0.498\\
    \multicolumn{1}{c|}{} & 720 & \textbf{0.363} & \textbf{0.384} & \underline{0.381} & \underline{0.401} & 0.400 & 0.403 & 0.424 & 0.416 & 0.426 & 0.417 & 0.474 & 0.449 & 0.436 & 0.440 & 0.433 & 0.427 & 0.467 & 0.442 & 0.712 & 0.614 & 1.913 & 1.166 & 0.589 & 0.557\\ \cline{2-26}
    \multicolumn{1}{c|}{} & Avg & \textbf{0.257} & \textbf{0.317} & \underline{0.263} & \underline{0.323} & 0.284 & 0.332 & 0.305 & 0.344 & 0.293 & 0.335 & 0.320 & 0.353 & 0.316 & 0.368 & 0.296 & 0.343 & 0.336 & 0.373 & 0.463 & 0.488 & 1.342 & 0.930 & 0.447 & 0.487\\

    \shline
    \end{tabular}
    }
\label{table7}
\end{table*}

\begin{table*}[h]
    \centering
    \small
    \tabcolsep=0.8mm
    \caption{Full results of few-shot learning on 5\% training data. The input sequence length is set to 512. The predictive lengths are set to $\{96, 192, 336, 720\}$. '-' means that 5\% data is not sufficient to continue a training set. We bold the best performance and underline the second best performance.}
    \vspace{-1.5em}
    \resizebox{\textwidth}{!}{
    \begin{tabular}{cc|cc|cc|cc|cc|cc||cc|cc||cc|cc|cc|cc|cc}
    \shline
    \multicolumn{2}{c}{\multirow{3}{*}{Method}} & \multicolumn{10}{c||}{\textbf{\textit{LLM-based}}} & \multicolumn{4}{c||}{\textbf{\textit{Non-Transformer-based}}} & \multicolumn{10}{c}{\textbf{\textit{Transformer-based}}} \\ \cline{3-26}
    \multicolumn{2}{c}{} & \multicolumn{2}{c|}{FiCoTS} & \multicolumn{2}{c|}{TimeVLM} & \multicolumn{2}{c|}{S$^{2}$IP-LLM} & \multicolumn{2}{c|}{TimeLLM} & \multicolumn{2}{c||}{GPT4TS} & \multicolumn{2}{c|}{TimesNet} & \multicolumn{2}{c||}{DLinear} & \multicolumn{2}{c|}{PatchTST} & \multicolumn{2}{c|}{iTransformer} & \multicolumn{2}{c|}{FEDformer} & \multicolumn{2}{c|}{Autoformer} & \multicolumn{2}{c}{ETSformer} \\ \cline{3-26} 
    \multicolumn{2}{c}{} & MSE & MAE & MSE & MAE & MSE & MAE & MSE & MAE & MSE & MAE & MSE & MAE & MSE & MAE & MSE & MAE & MSE & MAE & MSE & MAE & MSE & MAE & MSE & MAE\\
    \hline \hline
    \multicolumn{1}{c|}{\multirow{5}{*}{\rotatebox{90}{ETTh1}}} & 96 & \textbf{0.405} & \textbf{0.425} & \underline{0.417} & \underline{0.435} & 0.500 & 0.493 & 0.518  & 0.498 & 0.543 & 0.506 & 0.892 & 0.625 & 0.547 & 0.503 & 0.557 & 0.519 & 0.808 & 0.610 & 0.593 & 0.529 & 0.681 & 0.570 & 0.169 & 0.832 \\
    \multicolumn{1}{c|}{} & 192 & \textbf{0.428} & \textbf{0.440} & \underline{0.450} & \underline{0.458} & 0.690 & 0.539 & 0.702 & 0.547 & 0.748 & 0.580 & 0.940 & 0.665 & 0.720 & 0.604 & 0.711 & 0.570 & 0.928 & 0.658 & 0.652 & 0.563 & 0.725 & 0.602 & 1.221 & 0.853 \\
    \multicolumn{1}{c|}{} & 336 & \textbf{0.444} & \textbf{0.455} & \underline{0.460} & \underline{0.465} & 0.761 & 0.620 & 0.725 & 0.603 & 0.754 & 0.595 & 0.945 & 0.653 & 0.984 & 0.727 & 0.816 & 0.619 & 1.475 & 0.861 & 0.731 & 0.594 & 0.761 & 0.624 & 1.179 & 0.832\\
    \multicolumn{1}{c|}{} & 720 & - & - & - & - & - & - & - & - & - & - & - & - & - & - & - & - & - & - & - & - & - & - & - & - \\ \cline{2-26}
    \multicolumn{1}{c|}{} & Avg & \textbf{0.426} & \textbf{0.440} & \underline{0.442} & \underline{0.453} & 0.650 & 0.550 & 0.648 & 0.549 & 0.681 & 0.560 & 0.925 & 0.647 & 0.750 & 0.611 & 0.694 & 0.569 & 1.070 & 0.710 & 0.658 & 0.562 & 0.722 & 0.598 & 1.189 & 0.839 \\
    \hline \hline
    \multicolumn{1}{c|}{\multirow{5}{*}{\rotatebox{90}{ETTh2}}} & 96 & \textbf{0.292} & \textbf{0.355} & \underline{0.302} & \underline{0.365} & 0.363 & 0.409 & 0.384 & 0.420 & 0.376 & 0.421 & 0.409 & 0.420 & 0.442 & 0.456 & 0.401 & 0.421 & 0.397 & 0.427 & 0.390 & 0.424 & 0.428 & 0.468 & 0.678 & 0.619 \\
    \multicolumn{1}{c|}{} & 192 & \textbf{0.348} & \textbf{0.390} & \underline{0.361} & \underline{0.406} & 0.375 & 0.411 & 0.394 & 0.424 & 0.418 & 0.441 & 0.483 & 0.464 & 0.617 & 0.542 & 0.452 & 0.455 & 0.438 & 0.445 & 0.457 & 0.465 & 0.496 & 0.504 & 0.845 & 0.697 \\
    \multicolumn{1}{c|}{} & 336 & \textbf{0.370} & \textbf{0.413} & \underline{0.398} & \underline{0.434} & 0.403 & 0.421 & 0.416 & 0.433 & 0.408 & 0.439 & 0.499 & 0.479 & 1.424 & 0.849 & 0.464 & 0.469 & 0.631 & 0.553 & 0.477 & 0.483 & 0.486 & 0.496 & 0.905 & 0.727 \\
    \multicolumn{1}{c|}{} & 720 & - & - & - & - & - & - & - & - & - & - & - & - & - & - & - & - & - & - & - & - & - & - & - & - \\ \cline{2-26}
    \multicolumn{1}{c|}{} & Avg & \textbf{0.337} & \textbf{0.382} & \underline{0.354} & \underline{0.402} & 0.380 & 0.413 & 0.398 & 0.426 & 0.400 & 0.433 & 0.439 & 0.448 & 0.694 & 0.577 & 0.827 & 0.615 & 0.488 & 0.475 & 0.463 & 0.454 & 0.441 & 0.457 & 0.809 & 0.681 \\
    \hline \hline
    \multicolumn{1}{c|}{\multirow{5}{*}{\rotatebox{90}{ETTm1}}} & 96 & \textbf{0.308} & \textbf{0.351} & \underline{0.314} & \underline{0.357} & 0.357 & 0.390 & 0.422 & 0.424 & 0.386 & 0.405 & 0.606 & 0.518 & 0.332 & 0.374 & 0.399 & 0.414 & 0.589 & 0.510 & 0.628 & 0.544 & 0.726 & 0.578 & 1.031 & 0.747 \\
    \multicolumn{1}{c|}{} & 192 & \textbf{0.342} & \textbf{0.372} & \underline{0.343} & \underline{0.373} & 0.432 & 0.434 & 0.448 & 0.440 & 0.440 & 0.438 & 0.681 & 0.539 & 0.358 & 0.390 & 0.441 & 0.436 & 0.703 & 0.565 & 0.666 & 0.566 & 0.750 & 0.591 & 1.087 & 0.766 \\
    \multicolumn{1}{c|}{} & 336 & \textbf{0.370} & \textbf{0.387} & \underline{0.373} & \underline{0.391} & 0.440 & 0.442 & 0.452 & 0.447 & 0.485 & 0.459 & 0.786 & 0.597 & 0.402 & 0.416 & 0.499 & 0.467 & \textbf{0.898} & 0.641 & 0.807 & 0.628 & 0.851 & 0.659 & 1.138 & 0.787\\
    \multicolumn{1}{c|}{} & 720 & \textbf{0.424} & \textbf{0.419} & \underline{0.425} & \underline{0.420} & 0.593 & 0.521 & 0.585 & 0.491 & 0.577 & 0.499 & 0.796 & 0.593 & 0.511 & 0.489 & 0.767 & 0.587 & 0.948 & 0.671 & 0.822 & 0.633 & 0.857 & 0.655 & 1.245 & 0.831 \\ \cline{2-26}
    \multicolumn{1}{c|}{} & Avg & \textbf{0.361} & \textbf{0.382} & \underline{0.364} & \underline{0.385} & 0.455 & 0.446 & 0.477 & 0.451 & 0.472 & 0.450 & 0.717 & 0.561 & 0.400 & 0.417 & 0.526 & 0.476 & 0.784 & 0.596 & 0.730 & 0.592 & 0.796 & 0.620 & 1.125 & 0.782 \\
    \hline \hline
    \multicolumn{1}{c|}{\multirow{5}{*}{\rotatebox{90}{ETTm2}}} & 96 & \textbf{0.166} & \textbf{0.257} & \underline{0.169} & \underline{0.260} & 0.197 & 0.278 & 0.205 & 0.277 & 0.199 & 0.280 & 0.220 & 0.299 & 0.236 & 0.326 & 0.206 & 0.288 & 0.265 & 0.339 & 0.229 & 0.320 & 0.232 & 0.322 & 0.404 & 0.485 \\
    \multicolumn{1}{c|}{} & 192 & \textbf{0.223} & \textbf{0.296} & \underline{0.224} & \underline{0.298} & 0.254 & 0.322 & 0.267 & 0.336 & 0.256 & 0.316 & 0.311 & 0.361 & 0.306 & 0.373 & 0.264 & 0.324 & 0.310 & 0.362 & 0.394 & 0.361 & 0.291 & 0.357 & 0.479 & 0.521 \\
    \multicolumn{1}{c|}{} & 336 & \textbf{0.277} & \textbf{0.330} & \underline{0.282} & \underline{0.338} & 0.315 & 0.350 & 0.309 & 0.347 & 0.318 & 0.353 & 0.338 & 0.366 & 0.380 & 0.423 & 0.334 & 0.367 & 0.373 & 0.399 & 0.378 & 0.427 & 0.478 & 0.517 & 0.552 & 0.555 \\
    \multicolumn{1}{c|}{} & 720 & \textbf{0.364} & \textbf{0.385} & \underline{0.375} & \underline{0.397} & 0.421 & 0.421 & 0.448 & 0.432 & 0.460 & 0.436 & 0.509 & 0.465 & 0.674 & 0.583 & 0.454 & 0.432 & 0.478 & 0.454 & 0.523 & 0.510 & 0.553 & 0.538 & 0.701 & 0.627 \\ \cline{2-26}
    \multicolumn{1}{c|}{} & Avg & \textbf{0.257} & \textbf{0.317} & \underline{0.262} & \underline{0.323} & 0.296 & 0.342 & 0.307 & 0.348 & 0.308 & 0.346 & 0.344 & 0.372 & 0.399 & 0.426 & 0.314 & 0.352 & 0.356 & 0.388 & 0.381 & 0.404 & 0.388 & 0.433 & 0.534 & 0.547 \\

    \shline
    \end{tabular}
    }
\label{table8}
\end{table*}

\section{Full Long-term Forecasting Results}
\label{C}
The full long-term forecasting results are shown in Table \ref{figure6}.
FiCoTS outperforms all the baselines in most scenarios. It is worth mentioning that compared with TimeVLM, the state-of-art LLM-as-Enhancer model, FiCoTS achieves an average improvement of  2.0\% in MSE on all datasets. In addition, it outperforms the very recent LLM-as-Predictor models, S$^{2}$IP-LLM, Time-LLM and GPT4TS, by 2.9\%, 5.1\% and 4.7\%.

\section{Full Few-shot Forecasting Results}
\label{D}
The full few-shot forecasting results on 10\% and 5\% training data are shown in Table \ref{table7} and Table \ref{table8}, respectively. For few-shot forecasting using 10\% data, FiCoTS significantly surpasses baseline. Notably,  it out performs TimeVLM, S$^{2}$IP-LLM, Time-LLM and GPT4TS by 2.7\%, 21.8\&, 31.8\% and 21.9\%. For few-shot forecasting using 5\% data, FiCoTS achieves improvement of 3.0\%, 22.5\%, 24.6\% and 25.8\%.

\end{document}